\documentclass[11pt]{article}

\usepackage[preprint]{acl}
\usepackage{multirow}
\usepackage{booktabs}
\usepackage[table]{xcolor}
\usepackage{times}
\usepackage{latexsym}
\usepackage{makecell}
\usepackage{diagbox}

\usepackage[T1]{fontenc}
\usepackage[utf8]{inputenc}
\usepackage[english]{babel}

\usepackage{microtype}

\usepackage{inconsolata}

\usepackage{graphicx}

\usepackage{amsmath,amsfonts,bm}

\def\eqref#1{equation~\ref{#1}}
\def\1{\bm{1}}

\DeclareMathAlphabet{\mathsfit}{\encodingdefault}{\sfdefault}{m}{sl}
\SetMathAlphabet{\mathsfit}{bold}{\encodingdefault}{\sfdefault}{bx}{n}

\usepackage{url}
\usepackage{hyperref}

\usepackage{enumitem}
\usepackage{xspace}
\usepackage{algorithm}
\usepackage{algpseudocode}
\algrenewcommand\algorithmiccomment[1]{\hfill $\triangleright$ #1}
\usepackage{amsthm}
\usepackage{amsmath}
\usepackage{bm}
\usepackage{mathtools}
\usepackage{tabularx}
\usepackage{mdwlist}
\usepackage{subcaption}
\usepackage{tikz}
\usepackage{array}
\newcolumntype{C}[1]{>{\centering\arraybackslash}p{#1}}
\usepackage{ragged2e}
\usetikzlibrary{shapes,arrows.meta,positioning}

\usepackage{siunitx}
\usepackage{listings}
\lstdefinestyle{jsonstyle}{
  basicstyle=\ttfamily\footnotesize,
  morestring=[b]",
  showstringspaces=false,
  breaklines=true,
  breakatwhitespace=true,
  columns=fullflexible,
  keepspaces=true,
  frame=single,
  framesep=5pt,
  xleftmargin=5pt,
  xrightmargin=5pt,
  aboveskip=6pt,
  belowskip=2pt,
}

\newtheorem*{definition}{Definition}

\makeatletter
\renewcommand{\@listi}{\leftmargin\leftmargini \topsep 4pt \parsep 0pt \itemsep 1pt}
\let\@listI\@listi
\makeatother

\newcommand{\ul}[1]{\underline{#1}}
\newcommand{\hide}[1]{}

\newcommand{\method}{\textsc{Archive}\xspace}
\newcommand{\methodfull}{\ul{A}mbiguity \ul{R}ecognition via \ul{C}ascaded \ul{H}ypothesis \ul{I}nspection and Conflict \ul{VE}rification}

\newcommand{\kdamb}{KDAmb\xspace}
\newcommand{\sdamb}{SDAmb\xspace}
\newcommand{\closed}{Closed\xspace}
\newcommand{\open}{Open\xspace}
\newcommand{\mkdamb}{\mathrm{KDAmb}}
\newcommand{\mclosed}{\mathrm{Closed}}
\newcommand{\mopen}{\mathrm{Open}}

\newcommand{\impfonea}{10.4\%\xspace}
\newcommand{\impfoneu}{21.6\%\xspace}
\newcommand{\faster}{$16\times$\xspace}

\newcommand{\Famb}{F1-amb\xspace}
\newcommand{\Funamb}{F1-unamb\xspace}

\newcommand{\dataset}{\textsc{QuireQA}\xspace}
\newcommand{\datasetfull}{\ul{Q}ueries with \ul{U}nstructured and \ul{I}ncomplete \ul{R}eal-world \ul{E}xpressions}
\newcommand{\datasetsize}{4{,}703}
\newcommand{\ambtounamb}{206}
\newcommand{\contextadded}{3{,}299}
\newcommand{\orcasreview}{6{,}336}
\newcommand{\orcastotal}{10.4M}
\newcommand{\orcascandidates}{124K}
\newcommand{\orcasconfirmed}{1{,}511}

\title{Diversity is Not Ambiguity: Toward Accurate and Efficient\\ Ambiguity Detection for Open-Domain QA}

\author{
  Jiwon Lee \quad
  Yong-chan Park \quad
  Jungin Hong \quad
  U Kang\\
  Seoul National University\\
  \texttt{\{torijwl77,wjdakf3948,junginh,ukang\}@snu.ac.kr}
}

\begin{document}
\maketitle
\setlength{\abovedisplayskip}{8pt}
\setlength{\belowdisplayskip}{8pt}

%The abstract paragraph should be indented 1/2~inch (3~picas) on both left and
%right-hand margins. Use 10~point type, with a vertical spacing of 11~points.
%The word \textsc{Abstract} must be centered, in small caps, and in point size 12. Two
%line spaces precede the abstract. The abstract must be limited to one
%paragraph.

\begin{abstract}
How can question answering (QA) systems determine whether a query is ambiguous?
Ambiguity detection is essential in open-domain QA,
as misclassification leads to either answering the wrong interpretation or unnecessary clarification.
However, existing methods conflate answer diversity with ambiguity,
leading to inaccurate predictions.
They also process queries uniformly, resulting in wasteful computation.
We propose \method (\methodfull),
an accurate and efficient framework that detects ambiguity via \emph{logical conflict}:
a query is ambiguous when its valid answers cannot all be true
under a single interpretation.
\method combines a lightweight early-exit encoder for surface-detectable cases
with a conflict reasoning module that models logical relations among answers,
reinforced by an invariance objective for robustness to noisy answer sets.
We present \dataset, a \datasetsize{}-query benchmark
spanning factoid, non-factoid, and ill-formed queries.
Experiments show \method outperforms competitors,
improving \Famb by up to \impfonea and \Funamb by up to \impfoneu,
while operating \faster faster than the best competitor. 
\end{abstract}

\section{Introduction}
\label{sec:intro}

\begin{figure}[t]
	\centering
	\includegraphics[width=\columnwidth]{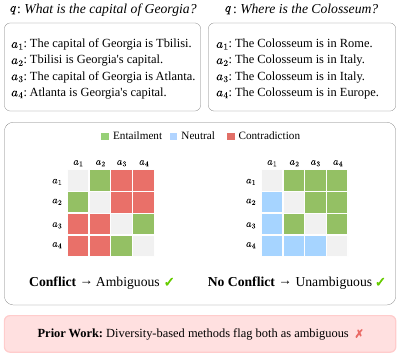}
	\caption{
    \textbf{Diversity is not ambiguity.}
    Both queries yield diverse answers, but only the first reflects genuinely conflicting interpretations.
    \method distinguishes them by testing for logical conflict rather than answer diversity.
	}
\label{fig:crown_jewel}
\end{figure}

%--------------------------------------------------------------

%%%%% 1. Problem Definition & Importance %%%%%
\textit{Given a user query $q$ in an open-domain question answering (QA) system,
how can we accurately and efficiently determine whether $q$ is ambiguous?}
Ambiguity detection is critical because users often issue underspecified queries
when they lack domain knowledge~\cite{min2020ambigqa}.
A QA system must either \emph{answer directly} when the query is unambiguous
or \emph{ask a clarifying question} when multiple interpretations exist~\cite{sun-etal-2023-answering}.
Missed ambiguity leads systems to commit to a single interpretation,
producing incorrect or misleading answers~\cite{liao2023proactive}.
Conversely, false positives trigger unnecessary clarification,
degrading interaction quality~\cite{kuhn2022clam}.

%%%%% 2. Challenges & Limitations of Prior Work %%%%%
Although LLMs are capable of generating clarifying questions,
the key challenge is deciding \emph{when} to do so~\cite{braslavski2017what}.
Existing approaches detect ambiguity through sets of candidate answers,
treating answer diversity as a signal for ambiguity~\cite{zhang2025clarify,shi2025trustnlp,
min2020ambigqa,lee2023asking,sun-etal-2023-answering}.
However, this view cannot distinguish
\emph{incompatible answers that arise from different interpretations}
from \emph{compatible answers that describe the same fact},
as both manifest as answer diversity.
For example, ``What is the capital of Georgia?'' yields incompatible answers (Tbilisi vs.\ Atlanta),
whereas ``Where is the Colosseum?'' yields diverse but compatible responses (Rome, Italy, Europe).
Prior methods label both as ambiguous due to answer diversity,
even though only the first exhibits genuine conflict (Figure~\ref{fig:crown_jewel}).
As a result, they generate unnecessary clarifying questions
for unambiguous queries (e.g., ``Do you mean the city or the country?'')
and incur unnecessary computation by routing every query through costly LLM pipelines.

%--------------------------------------------------------------
\begin{table*}[t]
\centering
\small
\caption{
    \textbf{Coverage gaps and non-exclusive labels in prior taxonomies.}
    Each row shows how a taxonomy labels four canonical query types;
    only \method provides a mutually exclusive assignment for every query.
}
\definecolor{naGray}{HTML}{E5E7EB}
\definecolor{overlapRed}{HTML}{FFCBC8}
\definecolor{archiveGreen}{HTML}{D6EDCB}
\newcolumntype{T}{>{\centering\arraybackslash}m{2.9cm}}
\newcolumntype{Q}{>{\centering\arraybackslash}m{2.9cm}}
\begin{tabular}{TQQQQ}
\toprule
\diagbox[width=2.9cm]{Taxonomy}{Query} & \makecell[c]{\emph{``what is the''}} & \makecell[c]{\emph{``What is the capital}\\\emph{of Georgia?''}} & \makecell[c]{\emph{``Where is the}\\\emph{Colosseum?''}} & \makecell[c]{\emph{``How should I}\\\emph{organize my desk?''}} \\
\midrule
\citet{zhang2024clamber} & \cellcolor{naGray}\textit{N/A} & Lexical & \cellcolor{naGray}\textit{N/A} & \cellcolor{overlapRed} What, Whom \\
\citet{tang2025clarifying} & \cellcolor{naGray}\textit{N/A} & \cellcolor{overlapRed} Semantic, Specify & \cellcolor{naGray}\textit{N/A} & Specify \\
\citet{tanjim2025detecting} & Syntactic & \cellcolor{naGray}\textit{N/A} & \cellcolor{naGray}\textit{N/A} & \cellcolor{naGray}\textit{N/A} \\
\midrule
\rowcolor{archiveGreen}
\shortstack[l]{\textbf{\method}\\\textbf{(proposed)}} & Surface-detectable ambiguous & Knowledge-dependent ambiguous & Closed unambiguous & Open unambiguous \\
\bottomrule
\end{tabular}

\vspace{4pt}
{\footnotesize
\textcolor{naGray}{\rule{8pt}{8pt}}\,No applicable class \quad
\textcolor{overlapRed}{\rule{8pt}{8pt}}\,Overlapping labels \quad
\textcolor{archiveGreen}{\rule{8pt}{8pt}}\,Mutually exclusive classes in \method}
\label{tab:taxonomy_comparison}
\end{table*}

\begin{figure}[t]
\centering
\includegraphics[width=\columnwidth]{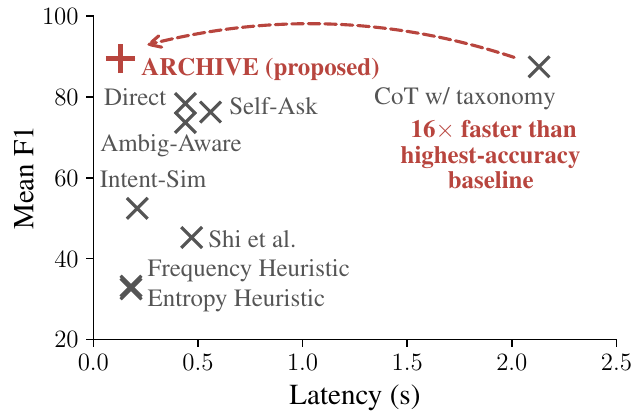}
\caption{
\textbf{\method achieves the highest mean F1 at the lowest latency on Qwen2.5-14B.}
Mean F1 averages \Famb and \Funamb.
}
\label{fig:crown_jewel_panels}
\end{figure}

%--------------------------------------------------------------

%%%%% 3. Main Contribution & Main Idea %%%%%
To address these limitations, we propose \method (\methodfull),
an accurate and efficient framework that redefines ambiguity detection through \emph{logical conflict}:
a query is ambiguous if and only if its valid answers cannot all be true under a single interpretation.
Unlike diversity-based heuristics, this formulation directly captures the underlying cause of ambiguity
and avoids conflating benign variation with genuine uncertainty.
Based on this formulation, \method adopts a cascaded architecture.
A lightweight encoder first resolves surface-detectable cases via early exit,
reducing unnecessary computation.
For remaining queries, \method generates candidate answers
and models logical relations among them to detect conflict.
To handle noisy or incomplete generations,
we introduce an invariance objective that stabilizes predictions
under perturbations of the candidate set.
Our contributions are as follows:
{\setlength{\leftmargini}{1em}
\setlength{\partopsep}{0pt}
\begin{itemize*}
    \item \textbf{Ambiguity via logical conflict.}
    We redefine ambiguity as logical conflict among candidate answers
    and present a mutually exclusive four-class taxonomy based on this formulation.
    \item \textbf{Accurate and efficient detection.}
    \method combines a lightweight early-exit encoder
    with a conflict reasoning module,
    reinforced by an invariance objective for robustness to noisy LLM generations.
    As shown in Figure~\ref{fig:crown_jewel_panels},
    it operates \faster faster than the highest-accuracy baseline.
    \item \textbf{Benchmark and experiments.}
    We construct \dataset, a \datasetsize{}-query benchmark with annotated context for factoid queries.
    \method outperforms nine baselines,
    improving \Famb by up to \impfonea and \Funamb by up to \impfoneu.
\end{itemize*}}

\noindent Code and data are available at: \url{https://github.com/snudatalab/ARCHIVE}.

\section{Related Work}
\label{sec:related}

\subsection{Prior Taxonomies of Ambiguity}
\label{sec:related_taxonomy}
Prior taxonomies enumerate categories of ambiguity without providing
an operational criterion for distinguishing ambiguous from unambiguous queries.
Earlier works categorized ambiguity types within specific QA datasets
without targeting detection~\cite{min2020ambigqa,guo2021abgcoqa,amplayo2022queryrefinementpromptsclosedbook}.
Three recent taxonomies address ambiguity detection:
\citet{zhang2024clamber} define eight categories across linguistic, epistemic, and aleatoric dimensions;
\citet{tang2025clarifying} propose three action-oriented classes to guide LLM response strategies,
explicitly aiming for a mutually exclusive partition;
\citet{tanjim2025detecting} categorize linguistic defects in conversational QA logs.
However, without a unifying criterion, these frameworks admit overlap:
a single query may satisfy multiple categories
(Table~\ref{tab:taxonomy_comparison}), leaving annotators with no principled way
to choose among labels.

\begin{figure}[t]
\centering
\includegraphics[width=\columnwidth]{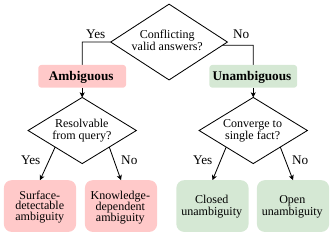}
\caption{
\textbf{Proposed taxonomy of query ambiguity.}
}
\label{fig:taxonomy}
\end{figure}

A second limitation is coverage.
\citet{zhang2024clamber} and \citet{tang2025clarifying} assume well-formed input
and do not account for surface defects, which are common in real query logs~\cite{craswell2020orcas}.
\citet{tanjim2025detecting} address this gap
but are scoped to enterprise data
and do not cover open-domain knowledge ambiguity.
These limitations motivate a different organizing principle:
rather than enumerating ambiguity categories,
we define ambiguity via \emph{logical conflict}---whether candidate answers
can be jointly true under a single interpretation of the query---and separate surface-detectable cases as a distinct class,
forming a mutually exclusive partition
(Figure~\ref{fig:taxonomy}, Section~\ref{sec:taxonomy}).

%--------------------------------------------------------------
\subsection{Ambiguity Detection}
\label{sec:related_detection}

Directly prompting an LLM to detect ambiguity is unreliable even for strong models~\cite{zhang2024clamber}.
Existing methods address this through indirect signals, but exhibit three main limitations.
First, these methods route every query through the same generation pipeline,
even when ambiguity is detectable from surface form alone.
This incurs unnecessary cost on trivially resolvable cases.

Second, diversity is an unreliable proxy.
A common approach estimates ambiguity from the diversity of generated outputs~\cite{cole2023},
for example via entropy over sampled answers~\cite{shi2025trustnlp, kuhn2023}
or LLM-simulated user responses~\cite{zhang2025clarify}.
However, this conflates \emph{aleatoric uncertainty} (genuine input ambiguity)
with \emph{epistemic uncertainty} (gaps in model knowledge)~\cite{hou2024decomposing}.
Hallucinations can inflate diversity and produce false positives,
while insufficient knowledge can suppress it and cause false negatives.
Diversity does not distinguish incompatible answers from benign variation.
For instance, ``What is the capital of Georgia?'' yields incompatible answers (Tbilisi vs.\ Atlanta),
whereas ``How should I organize my desk?'' yields diverse yet compatible suggestions.
Yet both produce high-entropy answer sets.
\citet{shi2025trustnlp} reduce the epistemic component using oracle context,
but still rely on output diversity.

Third, some methods define ambiguity relative to a particular model.
\citet{kim2024apa} measure ambiguity through information gain
between a query and its rewrite under a specific LLM,
while \citet{zhang2024clamber} formalize epistemic misalignment
with respect to the LLM's knowledge state.
Although framed as a feature of model-specific ambiguity,
such formulations couple detection to a particular model's capabilities,
making predictions sensitive to model choice and future updates.

In contrast, our proposed \method defines ambiguity via \emph{logical conflict}
among valid answers, rather than output diversity.
It filters surface-detectable cases before generation,
derives candidate answers from evidence context,
and trains for robustness against fabricated distractors.

% Taxonomy Figure
%--------------------------------------------------------------

% Overview figure
\begin{figure*}[t]
\centering
\includegraphics[width=\textwidth]{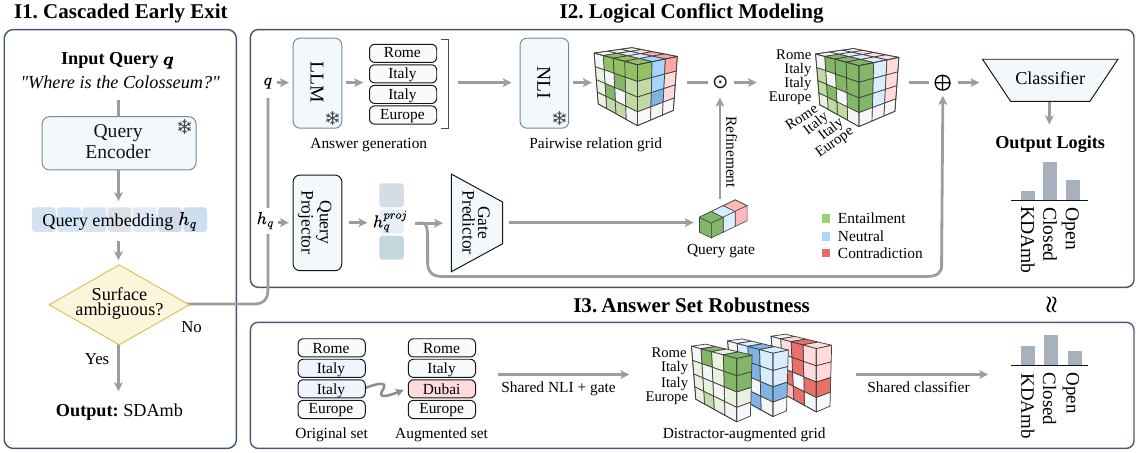}
\caption{
\textbf{Overview of \method.}
\method early-exits on surface-detectable ambiguous (\sdamb) queries, reducing unnecessary LLM calls (I1).
Remaining queries are classified by modeling pairwise logical relations among candidate answers (I2),
and robustness is enforced through prediction invariance after distractor augmentation~(I3).
}
\label{fig:overview}
\end{figure*}

%--------------------------------------------------------------

\section{Problem Formulation}
\label{sec:formulation}

We formalize ambiguity detection in two parts:
a taxonomy (Section~\ref{sec:taxonomy})
and a benchmark that applies this criterion to four query sources,
including queries that prior datasets label inconsistently under logical conflict (Section~\ref{sec:dataset}).

%--------------------------------------------------------------
\subsection{Proposed Taxonomy}
\label{sec:taxonomy}
Building on the limitations in Section~\ref{sec:related},
we define ambiguity by \emph{logical conflict} among valid answers.

\begin{definition}
A~query $q$ is \textbf{ambiguous} if its valid answers include $a_i, a_j$
that cannot be jointly satisfied under a single interpretation of $q$;
it is \mbox{\textbf{unambiguous}} if all valid answers are compatible.
\end{definition}

\paragraph{Taxonomy.}
As shown in Figure~\ref{fig:taxonomy},
queries are first classified as \emph{ambiguous} or \emph{unambiguous}
based on logical conflict, then divided into four classes:

{\setlength{\leftmargini}{1.5em}
\begin{enumerate*}
    \item \textbf{Surface-detectable ambiguity.}
    Queries whose plausible repairs or completions yield interpretations with conflicting answers
    (e.g., malformed syntax or unanchored references).

    \item \textbf{Knowledge-dependent ambiguity.}
    Well-formed queries admitting multiple interpretations with conflicting answers
    (e.g., ``capital of Georgia'': Tbilisi vs.\ Atlanta).

    \item \textbf{Closed unambiguity.}
    Answers converge to a single fact or bounded set
    (e.g., Rome, Italy).

    \item \textbf{Open unambiguity.}
    A single interpretation admits multiple compatible answers
    (e.g., diverse advice for organizing a desk).
\end{enumerate*}}

Surface-detectable ambiguity is prevalent in practice:
we apply heuristics (Appendix~\ref{app:orcas}) to the \orcastotal-query ORCAS corpus~\cite{craswell2020orcas}
and identify \orcascandidates{} candidate queries, of which \orcasconfirmed{}/\orcasreview{} reviewed cases 
are confirmed as surface-detectable ambiguous.

%--------------------------------------------------------------
\subsection{Dataset}
\label{sec:dataset}
Existing benchmarks focus on factoid queries with diversity-based ambiguity labels
that do not align with logical conflict.
We therefore construct \dataset\ (\datasetfull),
a benchmark of factoid and non-factoid queries from four sources,
re-annotated under our four-class taxonomy (Table~\ref{tab:dataset_stats}).
For example, AmbigNQ~\cite{min2020ambigqa} labels ``Where is the oldest bristlecone pine tree located?'' as ambiguous
because its answers span multiple geographic levels (peak, state, country),
yet these describe the same location at different granularities
and are closed unambiguous under our criterion.
Overall, \ambtounamb{} AmbigNQ queries flagged as ambiguous by previous methods
are unambiguous under our criterion.
To support future work on context-aware detection, we annotate and provide \contextadded{}
oracle context passages across 680 queries.
Details are in Appendix~\ref{app:dataset_app}.

\section{Proposed Method}
\label{sec:method}

We now present \method, a framework for accurate and efficient ambiguity detection in open-domain QA.
It addresses three key challenges:

{\setlength{\leftmargini}{7.2mm}
\setlength{\partopsep}{0pt}
\begin{itemize*}
\item[\textbf{C1.}] \textbf{Computational cost.}
Existing methods apply generation and reasoning to every query,
even when ambiguity is detectable from surface form alone.
How can we avoid unnecessary LLM inference while preserving accuracy?

\item[\textbf{C2.}] \textbf{Ambiguity vs.\ answer diversity.}
Prior work treats answer variability as a proxy for ambiguity.
However, diverse answers may be mutually compatible (e.g., open-ended queries),
while truly ambiguous queries exhibit conflicting interpretations.
How can we distinguish true ambiguity from benign diversity?

\item[\textbf{C3.}] \textbf{Unreliable LLM generations.}
LLMs may introduce hallucinated contradictions or omit valid alternative answers,
causing detectors to misestimate ambiguity.
How can predictions remain stable under such noise?
\end{itemize*}}

% Dataset Statistics
%--------------------------------------------------------------
\begin{table}[t]
\centering
\small
\caption{
\textbf{\dataset statistics.}
\sdamb/\kdamb: surface-detectable/knowledge-dependent ambiguous;
\closed/\open: closed/open unambiguous.
}
\label{tab:dataset_stats}
\setlength{\tabcolsep}{2pt}
\renewcommand{\arraystretch}{1.0}
\begin{tabular}{lrrrrrr}
\toprule
& \multicolumn{2}{c}{Ambiguous} & \multicolumn{2}{c}{Unambiguous} & \\
\cmidrule(lr){2-3} \cmidrule{4-5}
Source & \sdamb & \kdamb & \closed & \open & Total \\
\midrule
ORCAS~\shortcite{craswell2020orcas}  & 1{,}512  & 2 & 60 & 37 & 1{,}611 \\
AmbigNQ~\shortcite{min2020ambigqa}   & 0 & 414 & 1{,}254 & 11 & 1{,}679 \\
AmbER~\shortcite{chen-etal-2021-evaluating}     & 0  & 317 & 83 & 0 & 400 \\
Dolly-15K~\shortcite{DatabricksBlog2023DollyV2} & 0 & 4 & 61  & 948 & 1{,}013 \\
\midrule
Total & 1{,}512  & 737 & 1{,}458 & 996 & 4{,}703 \\
\bottomrule
\end{tabular}
\end{table}
%--------------------------------------------------------------
As illustrated in Figure~\ref{fig:overview}, \method tackles these challenges with the following main ideas:

{\setlength{\leftmargini}{7.2mm}
\setlength{\partopsep}{0pt}
\begin{itemize*}
\item[\textbf{I1.}] \textbf{Cascaded early exit (Section~\ref{sec:earlyexit}).}
A~lightweight encoder-based module detects surface-detectable ambiguity from the query,
bypassing LLM inference when~possible.

\item[\textbf{I2.}] \textbf{Logical conflict modeling (Section~\ref{sec:conflict}).}
For~remaining queries, we generate candidate answers and model pairwise relations
to detect logical conflict among them.

\item[\textbf{I3.}] \textbf{Answer set robustness (Section~\ref{sec:invariance}).}
\mbox{Distractor} augmentation with an invariance loss reduces sensitivity to hallucinated contradictions,
while context-grounded generation helps recover omitted valid answers.
\end{itemize*}}

The early-exit and conflict modeling modules are trained independently.
The former minimizes $\mathcal{L}_{\text{exit}}$,
whereas the latter optimizes
\[
\mathcal{L}_{\text{conflict}} = \mathcal{L}_{\text{CE}} + \lambda_{\text{inv}} \mathcal{L}_{\text{inv}},
\]
%$\mathcal{L}_{\text{conflict}} = \mathcal{L}_{\text{CE}} + \lambda_{\text{inv}} \mathcal{L}_{\text{inv}}$,
where $\mathcal{L}_{\text{CE}}$ supervises classification, $\mathcal{L}_{\text{inv}}$ enforces robustness to distractor perturbations,
and $\lambda_{\text{inv}}$ controls the invariance strength.
%--------------------------------------------------------------

\subsection{Cascaded Early Exit}
\label{sec:earlyexit}

The first stage of \method detects \emph{surface-detectable ambiguity} (\sdamb)
with a lightweight classifier on a frozen text encoder.
This reduces cost and latency by reserving LLM calls for queries that require deeper reasoning.
Given a query $q$, the encoder produces a mean-pooled query embedding $\mathbf{h}_q \in \mathbb{R}^d$.
A trainable linear classifier head with sigmoid activation estimates
$p = P(\text{\sdamb} \mid \mathbf{h}_q)$.
It is trained with binary cross-entropy:
\[
\mathcal{L}_{\text{exit}} = -\ell \log p - (1 - \ell) \log (1 - p),
\]
where $\ell \in \{0, 1\}$ denotes the \sdamb label.
At inference, queries with $p > \tau$ exit early and return the \sdamb label, where $\tau$ is a validation-tuned routing threshold.
Otherwise, \method forwards the query and its embedding to the conflict modeling module for deeper analysis.
%--------------------------------------------------------------
\subsection{Logical Conflict Modeling}
\label{sec:conflict}

For queries not handled by early exit,
the conflict modeling module predicts one of three labels:
knowledge-dependent ambiguity (\kdamb), closed unambiguity (\closed), or open unambiguity (\open).
The key insight is that ambiguity manifests as \emph{logical conflict}
among valid answers, not answer diversity.
The module thus generates candidate answers for a given query,
employs natural language inference (NLI) to organize their pairwise relations into a grid
that separates contradictions from compatible variations,
and classifies the query from a query-conditioned refinement of that grid.

%The module thus generates candidate answers for a query and employs
%
%natural language inference (NLI)
%to model their pairwise logical relations and distinguish contradictions from compatible variations.
%
%It organizes the relations into a grid, refines the grid via query-conditioned calibration,
%and classifies the query using the refined grid and query embedding.

\paragraph{Answer generation.}
For each query $q$, we first generate a set $\{a_1, \ldots, a_k\}$ of
$k{=}10$ candidate answers for conflict analysis.
We prompt an LLM five times to produce multiple answers covering different interpretations of $q$,
formatted as short declarative statements for reliable downstream NLI.
The top-$k$ most frequent unique answers constitute the final set.
If fewer than $k$ unique answers are produced, we pad the set by duplicating
existing entries in proportion to their frequency, ensuring a fixed-size grid.
Generation is conditioned on context when available, as described below.
See Appendices~\ref{app:impl_details} and~\ref{app:prompts} for generation details and prompts.

\paragraph{Pairwise relation grid.}
Rather than relying on indirect proxies,
we model the logical structure among candidate answers as pairwise NLI relations.
Given the answer set $\{a_1, \ldots, a_k\}$, we construct a relation grid $S \in \mathbb{R}^{k \times k \times 3}$.
For each ordered pair $(a_i, a_j)$ with $i \neq j$,
a frozen NLI model~\cite{laurer2022less} predicts a distribution over entailment ($E$), neutral ($N$), and contradiction ($C$),
indicating whether $a_i$ entails, is neutral to, or contradicts $a_j$:
\[
S_{i,j} = [p_E, p_N, p_C] \in [0,1]^3,\quad \sum_{\mathclap{r \in \{E,N,C\}}} p_r = 1.
\]

Thus, each channel of $S$ represents one relation type over all $k^2$ pairs.
Diagonal entries are zeroed because self-relations are uninformative.

\paragraph{
Query-conditioned relation refinement.}
The grid $S$ isolates answer-level conflict, the core signal of ambiguity under our definition.
However, its interpretation depends on the query.
Open unambiguous queries naturally produce neutral-dominated grids due to diverse compatible answers,
while closed unambiguous queries may occasionally appear similar when NLI fails to recognize equivalent facts.
To mitigate this noise, we apply a lightweight query gate
that recalibrates the entailment, neutral, and contradiction channels of $S$.
The query embedding $\mathbf{h}_q$ is first projected as
$\mathbf{h}_q^{\text{proj}} = f_{\text{proj}}(\mathbf{h}_q)$
and mapped to per-channel scaling factors:
\[
\mathbf{g} = 2\sigma(f_{\text{gate}}(\mathbf{h}_q^{\text{proj}})) \in \mathbb{R}^3,
\]
where $f_{\text{proj}}$ is a linear layer with ReLU activation,
$f_{\text{gate}}$ is a 2-layer MLP with ReLU activation,
and $\sigma$ denotes a sigmoid applied externally to the output of $f_{\text{gate}}$.
The factor of~$2$ centers $\mathbf{g}$ at the identity
($\mathbf{g}=\mathbf{1}$ leaves $S$ unchanged).
The query gate ${\mathbf{g} = [g_E, g_N, g_C]}$ acts as a channel-wise temperature on the relation logits:
\[
\tilde{S}_{i,j}
= \operatorname{softmax}_{r \in \{E,N,C\}}
\!\left( \log(S_{i,j} + \epsilon) \odot \mathbf{g} \right),
\]
where the softmax is applied for each pair $(i,j)$ over the three NLI channels,
$\odot$ is elementwise multiplication,
and $\epsilon$ ensures numerical stability.
$g_r > 1$ sharpens relation type $r$, whereas $g_r < 1$ suppresses it.
A single shared $\mathbf{g}$ is applied to all $k^2$ pairs for efficient query-dependent refinement;
pair-specific conflict remains encoded in the scores $S_{i,j}$.
Figure~\ref{fig:nli_grid_detail} in Appendix~\ref{app:nli_grids} shows examples of $\tilde{S}$.

\paragraph{Classification.}
Given the flattened grid $\text{vec}(\tilde{S}) \in \mathbb{R}^{3k^2}$
and projected query embedding $\mathbf{h}_q^{\text{proj}}$,
we generate class logits
\[
\hat{y} = f_{\text{cls}}\!\left( [\text{vec}(\tilde{S});\, \mathbf{h}_q^{\text{proj}}] \right) \in \mathbb{R}^3,
\]
where $f_{\text{cls}}$ is a 2-layer MLP with ReLU activation,
and $\text{vec}(\cdot)$ denotes vectorization.
Concatenating $\mathbf{h}_q^{\text{proj}}$ grounds grid patterns in query intent,
enabling finer distinctions (e.g., open vs.\ closed unambiguous) that channel-wise modulation alone cannot make.
The classifier is trained with cross-entropy:
\[
\mathcal{L}_{\text{CE}} = -\log \operatorname{softmax}(\hat{y})_{c^*},
\]
where $c^* \in \{\mkdamb, \mclosed, \mopen\}$ is the ground-truth class.

% where $c^* \in \{\mkdamb, \mclosed, \mopen\}$ denotes the ground-truth class:
% knowledge-dependent ambiguity, closed unambiguity, or open unambiguity.
%--------------------------------------------------------------
\subsection{Answer Set Robustness}
\label{sec:invariance}

The conflict modeling module depends on the candidate answer set,
but LLM generation may hallucinate unsupported answers or omit valid interpretations.
We address hallucination with distractor invariance and omission with context grounding.
%--------------------------------------------------------------
\begin{figure}[t]
\centering
\includegraphics[width=\linewidth]{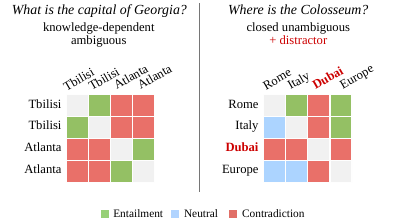}
\caption{
\textbf{Contradiction patterns distinguish genuine ambiguity from a distractor.}
Genuine ambiguity yields distributed contradictions (Left),
whereas a distractor yields isolated row/column contradictions (Right).
}
\label{fig:distractor_invariance}
\end{figure}
%--------------------------------------------------------------

\paragraph{Invariance to distractors.}
Hallucinated answers should not change the predicted ambiguity label.
We therefore train the classifier to be invariant to a synthetic distractor
inserted into the candidate answer set.
In \dataset, the LLM always produces fewer than $k{=}10$ unique answers.
Thus, at least one duplicate answer is always present in $\{a_1,\dots,a_k\}$,
which we replace with an LLM-generated distractor
(see Appendix~\ref{app:prompts} for the prompt).
This distractor matches the form of valid answers but is factually invalid,
contradicting the valid answers or available context.
As Figure~\ref{fig:distractor_invariance} shows, such distractors create \emph{pointwise} rather than distributed contradictions.
%Such a distractor induces \emph{pointwise} contradictions confined to its row and column,
%unlike the distributed pattern of genuine ambiguity (Figure~\ref{fig:distractor_invariance}).
We enforce invariance by minimizing the KL divergence between classifier logits on clean ($\hat{y}$) and augmented ($\hat{y}^{\text{aug}}$) inputs:
\[
\mathcal{L}_{\text{inv}} =
\operatorname{KL}\!\left(
\operatorname{softmax}(\hat{y}^{\text{aug}}) \;\|\;
\operatorname{softmax}(\hat{y})
\right).
\]

\paragraph{Context grounding.}
Distractor invariance handles hallucinated answers,
but cannot recover valid answers the LLM omits when generating without context.
We thus provide an oracle context passage---a relevant document supplied with the query---during answer generation.
This anchors candidate answers in external evidence and better covers the query's valid answer space.

%--------------------------------------------------------------

% Algorithm
\begin{algorithm}[t]
\caption{\method Inference}
\label{alg:archive}
\begin{algorithmic}[1]
\Require Query $q$, context $c$, and threshold $\tau$
\Ensure $y \in \{\text{SDAmb}, \text{KDAmb}, \text{Closed}, \text{Open}\}$

\State $\mathbf{h}_q, p \gets \mathrm{EarlyExit}(q)$ \Comment{surface check}
\If{$p > \tau$}
    \State \Return \sdamb \Comment{early exit}
\EndIf

\State $\{a_i\}_{i=1}^{k} \gets \mathrm{LLM}(q,c)$
\vspace{2pt}
\State $S \gets \mathrm{NLI}(a_1,\dots,a_k)$ \Comment{relation grid}
\State $\mathbf{g} \gets 2\sigma(f_{\mathrm{gate}}(f_{\mathrm{proj}}(\mathbf{h}_q)))$ \Comment{query gate}
\vspace{2pt}
\State $\tilde{S} \gets \mathrm{Refine}(S,\mathbf{g})$ \Comment{refinement}
\vspace{2pt}
\State $\hat{y} \gets f_{\mathrm{cls}}([\mathrm{vec}(\tilde{S}); f_{\mathrm{proj}}(\mathbf{h}_q)])$ \Comment{classification}
\State \Return $\arg\max \hat{y}$
\end{algorithmic}
\end{algorithm}

\subsection{Inference}
Algorithm~\ref{alg:archive} shows the inference pipeline.
The early-exit module first estimates the probability of surface-detectable ambiguity;
if $p>\tau$, the pipeline exits without LLM calls.
Otherwise, the conflict modeling module generates answers,
builds and refines an NLI relation grid,
and combines the grid with the query embedding to predict $\arg\max \hat{y}$.

% RQ1 Table
%--------------------------------------------------------------
\begin{table*}[!t]
\centering
\small
\caption{
\textbf{Defining ambiguity by logical conflict improves taxonomy utility.}
Rows vary only by taxonomy.
CQ/Ans BERTScore are computed on knowledge-dependent ambiguous and closed-unambiguous queries, respectively.
}
\renewcommand{\arraystretch}{0.9}
\newcolumntype{M}{>{\centering\arraybackslash}p{2.1cm}}
\begin{tabular*}{\textwidth}{@{\extracolsep{\fill}}llMMMM}
\toprule
Backbone & Taxonomy & \Famb & \Funamb & CQ BERTScore & Ans BERTScore \\
\midrule
\multirow{4}{*}{LLaMA-2-7B}
& \citet{zhang2024clamber} & 36.20 & \underline{28.57} & 82.08 & \underline{80.81} \\
& \citet{tang2025clarifying} & \underline{64.99} & \phantom{0}0.00 & \textbf{84.37} & 79.42 \\
& \citet{tanjim2025detecting} & 64.92 & \phantom{0}0.00 & \underline{82.51} & 78.68 \\
\cmidrule(l){2-6}
& \textbf{\method (proposed)} & \textbf{69.83} & \textbf{64.21} & \underline{83.69} & \textbf{83.97} \\
\midrule
\multirow{4}{*}{Qwen2.5-14B}
& \citet{zhang2024clamber} & \underline{69.58} & \underline{36.70} & 81.34 & 82.04 \\
& \citet{tang2025clarifying} & 65.82 & 17.06 & \underline{86.18} & 82.57 \\
& \citet{tanjim2025detecting} & 67.86 & 28.42 & 83.22 & \underline{83.01} \\
\cmidrule(l){2-6}
& \textbf{\method (proposed)} & \textbf{75.25} & \textbf{64.84} & \textbf{86.70} & \textbf{83.76} \\
\bottomrule
\end{tabular*}
\label{tab:taxonomy_eval}
\vspace{5pt}
\end{table*}

\section{Experiments}
\label{sec:experiments}

We evaluate \method across five key questions:

{\setlength{\leftmargini}{5.5mm}
\begin{itemize*}
\item[\textbf{Q1.}] \textbf{Taxonomy utility (Section~\ref{sec:exp_taxonomy}).}
Does the proposed taxonomy outperform existing ones in ambiguity detection and response generation?

\item[\textbf{Q2.}] \textbf{Binary ambiguity detection (Section~\ref{sec:exp_performance}).}
Does \method improve binary ambiguity detection accuracy over prior methods?

\item[\textbf{Q3.}] \textbf{Fine-grained detection (Section~\ref{sec:exp_4way}).}
Does \method accurately predict the four classes?

\item[\textbf{Q4.}] \textbf{Efficiency (Section~\ref{sec:exp_efficiency}).}
Does the cascade reduce inference cost while preserving accuracy?

\item[\textbf{Q5.}] \textbf{Ablation (Section~\ref{sec:exp_ablation}).}
How much does each module contribute to the overall performance?
\end{itemize*}}

\noindent
See Appendices~\ref{app:hyperparameter} and~\ref{app:exp_context} for additional analyses.
%Additional analyses on hyperparameter and context quality sensitivity are in
%Appendices~\ref{app:hyperparameter} and~\ref{app:exp_context}.

%--------------------------------------------------------------
\subsection{Experimental Setup}
\label{sec:exp_setup}

\paragraph{Dataset.}
We train and evaluate on \dataset, using 8:1:1 train/validation/test split.
For out-of-distribution evaluation, we additionally test on AmbigTriviaQA~\cite[Appendix~\ref{app:dataset_construction}]{kim2024apa}.

\paragraph{Backbones.}
The conflict modeling module is model-agnostic; we instantiate it with
LLaMA-2-7B~\cite{touvron2023llama2} and Qwen2.5-14B-Instruct~\cite{qwen2025qwen25}.
The early-exit module uses DeBERTa-v3-base~\cite{he2023debertav3}.
Pairwise relations are computed with the released DeBERTa-v3 NLI checkpoint
of \citet{laurer2022less}.
All backbone models are kept frozen.

\paragraph{Competitors.}
We compare with nine methods:
(1)~\emph{prompting}: Direct,
a simple QA prompt,
Ambig-Aware~\cite{kim2024apa} with ambiguity-specific instructions,
Self-Ask~\cite{amayuelas2024knowledge}, which generates an answer before judging ambiguity,
and CoT w/ taxonomy, adapted from \citet{tang2025clarifying} using our taxonomy;
(2)~\emph{diversity heuristics}: Frequency, Entropy, Intent-Sim~\cite{zhang2025clarify},
which infer ambiguity from answer diversity; and
(3)~\emph{trained models}: APA~\cite{kim2024apa} and \citet{shi2025trustnlp}.
Details are in Appendix~\ref{app:competitor_details}.

\paragraph{Metrics.}
We report per-class F1 (\Famb, \Funamb) for binary detection;
4-way classification uses macro- and per-class F1.
Generation quality is measured with BERTScore~\cite{zhang2020bertscoreevaluatingtextgeneration}
for answers and clarifying questions.
Latency is the mean end-to-end wall-clock time per query.
All results are averaged over 3 seeds.

%--------------------------------------------------------------
\subsection{Taxonomy Utility (Q1)}
\label{sec:exp_taxonomy}
We fix the prompt and vary only the taxonomy schema,
evaluating all queries with LLM generation conditioned on oracle context.
CQ BERTScore and Ans BERTScore measure response quality on
\kdamb and \closed queries, respectively---the classes with available references.
We construct pseudo-reference clarifying questions by adapting the clarification prompt of
\citet{kim2024apa} to each query and its gold disambiguation (details are in Appendix~\ref{app:taxonomy_exp}).
Table~\ref{tab:taxonomy_eval} shows that our taxonomy yields the strongest detection
and competitive response quality relative to
symptom-based~\cite{zhang2024clamber}, action-oriented~\cite{tang2025clarifying},
and linguistic~\cite{tanjim2025detecting} alternatives,
supporting logical conflict as a useful criterion.

%--------------------------------------------------------------
% RQ2 Table
\begin{table}[t]
\centering
\small
\caption{
	\textbf{\method achieves the highest F1 in binary ambiguity detection.}
	Factoid queries with no context are used, matching the setup of prior detection work.
}
\renewcommand{\arraystretch}{1.1}
\begin{tabular*}{\columnwidth}{@{}l@{\extracolsep{\fill}}rrrr@{}}
\toprule
\multirow{2}[3]{*}{{Method}} & \multicolumn{2}{c}{QuireQA} & \multicolumn{2}{c}{AmbigTriviaQA} \\
\cmidrule(lr){2-3} \cmidrule(lr){4-5}
& \makecell{F1-\\[-2pt]amb} & \makecell{F1-\\[-2pt]unamb} & \makecell{F1-\\[-2pt]amb} & \makecell{F1-\\[-2pt]unamb} \\
\midrule
\multicolumn{5}{@{}l}{Backbone: \textit{LLaMA-2-7B}} \\[-1pt]
\midrule[0.3pt]
Direct & 2.77 & 59.52 & 55.99 & 1.17 \\
Ambig-Aware & 49.70 & 0.00 & \underline{56.76} & 0.80 \\
Self-Ask & 33.98 & 64.04 & 54.61 & 56.54 \\
CoT w/ taxonomy & 49.85 & 4.01 & 49.14 & 57.80 \\
Frequency & 49.70 & 0.00 & 56.67 & 0.00 \\
Entropy & 48.80 & 32.14 & 39.72 & 30.87 \\
Intent-Sim & 41.84 & \underline{66.57} & 28.01 & 48.13 \\
APA & \underline{52.24} & 20.94 & 25.53 & \underline{60.06} \\
\midrule
\textbf{\method} & \textbf{58.96} & \textbf{85.57} & \textbf{60.55} & \textbf{66.58} \\
\midrule
\multicolumn{5}{@{}l}{Backbone: \textit{Qwen2.5-14B}} \\[-1pt]
\midrule[0.3pt]
Direct & 3.74 & 68.76 & 25.34 & 59.60 \\
Ambig-Aware & 45.32 & 69.18 & 65.75 & 61.98 \\
Self-Ask & 34.51 & \underline{72.43} & 65.54 & \underline{68.22} \\
CoT w/ taxonomy & 45.27 & 69.87 & 69.67 & 62.39 \\
Frequency & 49.70 & 0.00 & 66.67 & 0.00 \\
Entropy & 49.67 & 2.42 & 66.84 & 6.80 \\
Intent-Sim & 48.68 & 15.91 & \underline{71.08} & 59.67 \\
APA & \underline{50.88} & 64.05 & 53.95 & 67.22 \\
\midrule
\textbf{\method} & \textbf{61.31} & \textbf{85.66} & \textbf{79.19} & \textbf{69.04} \\
\bottomrule
\end{tabular*}
\label{tab:binary}
\end{table}

% RQ3 Table
\begin{table*}[!t]
\centering
\small
\caption{
\textbf{\method outperforms baselines in 4-way classification.}
We evaluate on all queries with the LLM conditioned on oracle context when available.
%
%SDAmb/KDAmb: surface-detectable/knowledge-dependent ambiguous; Closed/Open: closed/open unambiguous.
}
\begin{tabular*}{\textwidth}{@{}l@{\hspace{1.3em}}l@{\extracolsep{\fill}}c
@{\hspace{0.5em}}C{0.085\textwidth}
@{\hspace{0.5em}}C{0.085\textwidth}
@{\hspace{0.5em}}C{0.085\textwidth}
@{\hspace{0.5em}}C{0.085\textwidth}@{}}
\toprule
\multirow{2}[3]{*}{Backbone} & \multirow{2}[3]{*}{Method} & \multirow{2}[3]{*}{Macro-F1}
& \multicolumn{4}{c}{Per-Class F1} \\
\cmidrule(l{0.7em}r{0.1em}){4-7}
 & & & SDAmb & KDAmb & Closed & Open \\
\midrule
\multirow{2}{*}{LLaMA-2-7B} & CoT w/ taxonomy & 41.04 & 79.99 & 35.54 & \phantom{0}6.29 & 42.33 \\
 & \textbf{\method (proposed)} & \textbf{84.63} & \textbf{99.68} & \textbf{61.86} & \textbf{80.68} & \textbf{92.29} \\
\midrule
\multirow{2}{*}{Qwen2.5-14B} & CoT w/ taxonomy & 76.99 & 96.17 & 59.33 & 74.70 & 77.76 \\
 & \textbf{\method (proposed)} & \textbf{85.55} & \textbf{99.79} & \textbf{65.89} & \textbf{81.73} & \textbf{94.78} \\
\bottomrule
\end{tabular*}
\label{tab:4way}
\end{table*}
%--------------------------------------------------------------
\subsection{Binary Ambiguity Detection (Q2)}
\label{sec:exp_performance}
Table~\ref{tab:binary} compares \method against competitors
on factoid queries in a context-free setting, following prior work.
\citet{shi2025trustnlp} assume access to context and thus are omitted from this comparison.
First, \method achieves the highest \Famb and \Funamb
across both datasets and backbones.
This confirms that modeling logical conflict captures ambiguity more reliably than diversity proxies.
Baselines struggle to balance both metrics: direct prompting often collapses to a single class,
while the frequency heuristic yields zero \Funamb by counting paraphrases as separate answers.
%--------------------------------------------------------------
\subsection{4-Way Classification (Q3)}
\label{sec:exp_4way}
Table~\ref{tab:4way} reports 4-way classification using our taxonomy across all queries, with the LLM given oracle context.
\method outperforms CoT w/ taxonomy on macro-F1,
with near-perfect performance on \sdamb queries
and large gains on unambiguous classes (\closed, \open).
While \kdamb queries remain most challenging,
pairwise relation modeling still delivers consistent improvements,
highlighting its ability to capture logical conflict.
%--------------------------------------------------------------
% RQ4 Table
\begin{table}[t]
\centering
\small
\caption{
\textbf{\method achieves the highest F1 at the lowest latency.}
Results use Qwen2.5-14B on all queries, with oracle context when available.
}
\begin{tabular*}{\columnwidth}{@{}l@{\extracolsep{\fill}}cccc@{}}
\toprule
Method & \makecell{LLM\\[-2pt] Calls} & \makecell{Latency\\[-2pt] (s)} & \makecell{F1-\\[-2pt]amb} & \makecell{F1-\\[-2pt]unamb} \\
\midrule[0.3pt]
Direct & \textbf{1} & 0.44 & 68.55 & 78.76 \\
Ambig-Aware & \textbf{1} & 0.44 & 78.57 & 78.17 \\
CoT w/ taxonomy & \textbf{1} & 2.13 & \underline{86.72} & \underline{88.16} \\
Self-Ask & \underline{2} & 0.56 & 75.35 & 77.14 \\
Frequency & 10 & \underline{0.18} & 64.99 & \phantom{0}0.00 \\
Entropy & 10 & \underline{0.18} & 64.95 & \phantom{0}1.40 \\
Shi et al. & 10 & 0.47 & 44.78 & 45.65 \\
Intent-Sim & 11 & 0.21 & 65.31 & 39.53 \\
\midrule
\textbf{\method} & 3.34 & \textbf{0.13} & \textbf{89.11} & \textbf{89.87} \\
\bottomrule
\end{tabular*}
\label{tab:efficiency}
\end{table}
%--------------------------------------------------------------
\subsection{Efficiency Analysis (Q4)}
\label{sec:exp_efficiency}	
Table~\ref{tab:efficiency} compares inference cost and accuracy on all queries.
Filtering \sdamb queries through the early-exit module,
\method uses 3.34 LLM calls per query on average, versus 10 for heuristic and sampling-based baselines.
It achieves the lowest latency (0.13s) and highest accuracy,
showing that relation analysis outperforms brute-force sampling.
%--------------------------------------------------------------
% RQ 5 Table
\begin{table}[t]
\centering
\small
\caption{
\textbf{Each component of \method contributes.}
Results use Qwen2.5-14B on all queries, with oracle context when available.
}
\label{tab:ablation}
\begin{tabular*}{\columnwidth}{@{}l@{\extracolsep{\fill}}cccc@{}}
\toprule
Variant & \makecell{Macro-\\[-2pt]F1} & \makecell{F1-\\[-2pt]amb} & \makecell{F1-\\[-2pt]unamb} & \makecell{Latency\\[-2pt](s)} \\
\midrule
w/o routing & 53.34 & 30.78 & 68.63 & 0.16 \\
w/o $\mathcal{L}_{\text{inv}}$ & 83.46 & 87.99 & 86.12 & 0.13 \\
w/o $\tilde{S}$ ($\mathbf{h}_q$ only) & 80.67 & 87.55 & 85.45 & 0.14 \\
w/o gate & 82.22 & 86.89 & 87.94 & 0.15 \\
\method (full) & \textbf{85.55} & \textbf{89.11} & \textbf{89.87} & \textbf{0.13} \\
\bottomrule
\end{tabular*}
\end{table}
%--------------------------------------------------------------
\vspace*{-2em}
\subsection{Ablation Study (Q5)}
\label{sec:exp_ablation}
Table~\ref{tab:ablation} analyzes the contribution of each \method component.
Removing the early-exit module forces all queries through conflict modeling,
increasing latency because every query invokes the LLM and removing the only route for \sdamb queries.
Macro-F1 drops sharply, confirming early exit is a required routing component, not merely a speed optimization.
Removing $\mathcal{L}_{\text{inv}}$ reduces both F1 metrics.
Using only the query embedding $\mathbf{h}_q$ causes the largest macro-F1 drop among non-routing components,
showing pairwise relations are essential.
Removing the gate degrades \Famb the most, confirming the value of channel-wise recalibration.

\section{Conclusion}
\label{sec:conclusion}

We propose a four-class taxonomy based on \emph{logical conflict}
and construct \dataset, a benchmark for its evaluation.
We present \method, a framework for accurate and efficient ambiguity detection in open-domain QA.
\method combines an early-exit module for LLM-free detection of visibly ambiguous or malformed queries,
and a conflict modeling module for query-conditioned NLI over candidate answers.
Experiments show that \method improves \Famb by up to \impfonea and \Funamb by up to \impfoneu
with \faster faster inference than the highest-accuracy baseline.
These results illustrate that modeling ambiguity via structured logical conflict
outperforms diversity-based proxies in both detection quality and computational cost. 

\section*{Limitations}
\label{sec:limitations}
We acknowledge several limitations of this work.
First, \method is evaluated only on English-language queries;
extending the taxonomy and pipeline to multilingual settings remains an open direction.
Second, the conflict modeling module depends on a frozen LLM to generate candidate answers.
Although query-gated refinement and distractor augmentation mitigate hallucination noise,
a weak backbone can still limit performance.
Third, although \dataset aggregates multiple QA benchmarks,
it may not fully capture the distribution of queries in real-world deployments:
we exclude queries that are underspecified due to external dependencies.
Specifically, we filter out (1) temporally evolving queries whose answers change over time
(e.g., ``When will the next Olympics be held?'') and
(2) user-dependent queries (e.g., ``What's my IP address?'').
This filtering keeps ground-truth annotations stable and reproducible.
Finally, distinguishing knowledge-dependent ambiguity from closed unambiguity
depends on the NLI model's ability to recognize genuine contradictions among candidate answers.
Standard NLI training emphasizes textual entailment between sentence pairs,
but our task often requires entity-level knowledge to determine whether two candidate answers are genuinely incompatible.
When the NLI model lacks this knowledge, it may assign weak contradiction scores.
The query-conditioned gate has limited corrective capacity in these cases
because it rescales existing signals rather than introducing new knowledge.
Incorporating knowledge-aware or domain-adapted NLI models is a promising direction.

\section*{Ethical Considerations}
\label{sec:ethics}
All source datasets are publicly available,
contain no personally identifiable information,
and were vetted by their authors.
During curation, we additionally sample-checked 100 queries per source
for personally identifying information and offensive content; we found none.
There is an inherent risk of propagating social biases in those models
because \method relies on frozen LLMs to generate candidate answers;
we manually inspected generated outputs and found no harmful content,
though subtle biases may have been overlooked.
\method serves as a classification module rather than a user-facing generation system,
but ambiguity detectors could be misused to suppress legitimate user intent
and should be deployed with safeguards.

%\section*{Acknowledgments}

% Bibliography entries for the entire Anthology, followed by custom entries
%\bibliography{custom,anthology-overleaf-1,anthology-overleaf-2}

% Custom bibliography entries only
\bibliography{main}

\clearpage

\appendix
\begingroup
\setlength{\floatsep}{16pt plus 2pt minus 2pt}
\setlength{\textfloatsep}{16pt plus 2pt minus 2pt}
\setlength{\intextsep}{12pt plus 2pt minus 2pt}
\setlength{\dblfloatsep}{16pt plus 2pt minus 2pt}
\setlength{\dbltextfloatsep}{16pt plus 2pt minus 2pt}
\makeatletter
\renewcommand{\@listi}{\leftmargin\leftmargini \topsep 6pt \parsep 0pt \itemsep 3pt}
\let\@listI\@listi
\makeatother
%--------------------------------------------------------------
% Notation
%--------------------------------------------------------------
\section{Notation}
\label{app:notation}
Table~\ref{tab:notation} summarizes the paper's notation.

\begin{table}[t]
\centering
\small
\renewcommand{\arraystretch}{1.11}
\caption{Summary of notation.}
\begin{tabularx}{\columnwidth}{cX}
\toprule
\textbf{Symbol} & \textbf{Description} \\
\midrule
$q$ & User query \\
$c$ & Optional context passage \\
$y$ & Ambiguity label $\in \{$SDAmb, KDAmb, Closed, Open$\}$ \\
$\hat{y}$ & Predicted class logits \\
$\ell$ & Binary surface-detectable ambiguous label ($1$ if $q$ is SDAmb, else $0$) \\
$\tau$ & Early-exit routing threshold \\
$d$ & Encoder embedding dimension \\
$k$ & Size of the candidate answer set \\
$\mathbf{h}_q$ & Query representation $\in \mathbb{R}^d$ \\
$\mathbf{h}_q^{\text{proj}}$ & Projected query representation \\
$\mathbf{g}$ & Query-conditioned gate vector $\in \mathbb{R}^3$ \\
$S$ & Pairwise NLI relation grid $\in \mathbb{R}^{k \times k \times 3}$ \\
$\tilde{S}$ & Refined relation grid (query-gated) \\
$\mathcal{L}_{\text{exit}}$ & Binary cross-entropy loss for the \newline early-exit module \\
$\mathcal{L}_{\text{CE}}$ & Cross-entropy classification loss \\
$\mathcal{L}_{\text{inv}}$ & Invariance loss \\
$\lambda_{\text{inv}}$ & Weight for invariance loss \\
$\epsilon$ & Numerical stability constant \\
\bottomrule
\end{tabularx}
\label{tab:notation}
\end{table}

%--------------------------------------------------------------
% Dataset Construction
%--------------------------------------------------------------
\begin{table*}[!ht]
\centering
\small
\caption{Representative examples in \dataset by source and taxonomy class.}
\label{tab:dataset_examples}
\begin{tabular}{@{}>{\centering\arraybackslash}p{0.15\textwidth} >{\centering\arraybackslash}m{0.25\textwidth} m{0.55\textwidth}@{}}
\toprule
Source & Taxonomy Class & Example Query \& Details \\
\midrule
\multirow{5}{=}[-15pt]{\centering{ORCAS}}
& \multirow{3}{*}{\makecell{Surface-detectable\\ambiguous}} & \textbf{Query:} iuuui \\
& & \textbf{Query:} what is the \\
& & \textbf{Query:} who sings this \\
\cmidrule{2-3}
& Closed unambiguous & \textbf{Query:} what kind of word is and \newline \textbf{Answer:} a conjunction \\
\cmidrule{2-3}
& Open unambiguous & \textbf{Query:} what does asdfghjkl mean \newline \textbf{Answer:} Often used to represent random typing or keysmashing \dots \\
\midrule
\multirow{3}{=}[-20pt]{\centering{AmbigNQ}}
& \makecell{Knowledge-dependent\\ambiguous} & \textbf{Query:} When did the Simpsons first air on television? \newline \textbf{Answers:} April 19, 1987 (animated short) vs. December 17, 1989 (prime time show) \\
\cmidrule{2-3}
& Closed unambiguous & \textbf{Query:} What is the greek word for city state? \newline \textbf{Answer:} polis \\
\cmidrule{2-3}
& Open unambiguous & \textbf{Query:} What kind of meat is used for shabu shabu? \newline \textbf{Answer:} top sirloin, ribeye steak, Wagyu, \dots \\
\midrule
\multirow{2}{=}[-11pt]{\centering{AmbER}}
& \makecell{Knowledge-dependent\\ambiguous} & \textbf{Query:} What does Billy Preston play? \newline \textbf{Answers:} basketball (the athlete) vs. piano (the musician) \\
\cmidrule{2-3}
& Closed unambiguous & \textbf{Query:} Who directed The Godfather? \newline \textbf{Answer:} Francis Ford Coppola \\
\midrule
\multirow{2}{=}[-11pt]{\centering{Dolly-15K}}
& Open unambiguous & \textbf{Query:} How can I sleep well at night? \newline \textbf{Answer:} Here are some suggestions for a good night's rest \dots \\
\cmidrule{2-3}
& Closed unambiguous & \textbf{Query:} How many planets make up the Solar System? \newline \textbf{Answer:} 8 \\
\bottomrule
\end{tabular}
\end{table*}

% Open-answer generation prompt
\begin{table*}[t]
\centering
\small
\caption{\textbf{Prompt for generating representative answers for open unambiguous ORCAS queries.}
The prompt produces one valid answer rather than enumerating the full answer space.}
\label{tab:prompt_open_answers}
\begin{tabular}{@{}p{\textwidth}@{}}
\toprule
\makebox[\textwidth][c]{%
\begin{minipage}{0.95\textwidth}
\textbf{Task.}
Given an open-ended unambiguous query, generate a valid representative answer.

\textbf{Rules.}
\begin{enumerate}[nosep,leftmargin=1.4em,label=\arabic*.]
\item Provide one plausible answer that directly addresses the query.
\item If the query requires factual information, provide an answer that is broadly correct and verifiable; do not invent unsupported details.
\item Do not attempt to enumerate all possible valid answers.
\item Keep the answer concise: one sentence or a short list is preferred.
\item Return only valid JSON: \{"answer": "..."\}.
\end{enumerate}

Query: \texttt{\{query\}} \\
Answer:
\end{minipage}}
\\
\bottomrule
\end{tabular}
\end{table*}
%--------------------------------------------------------------
\section{QuireQA Construction}
\label{app:dataset_app}
We construct \dataset, a benchmark of factoid and non-factoid queries
under our four-class taxonomy.
\dataset draws from four sources:
ORCAS~\cite{craswell2020orcas} for web-search queries, including surface defects;
AmbigNQ~\cite{min2020ambigqa} for factoid queries derived from Natural Questions~\cite{kwiatkowski-etal-2019-natural} with crowdsourced disambiguations;
AmbER~\cite{chen-etal-2021-evaluating} for Wikipedia-grounded entity disambiguation queries;
and Dolly-15K~\cite{DatabricksBlog2023DollyV2} for human-written instructions with open-ended and closed-form unambiguous queries.
We annotate \contextadded{} oracle context passages from English Wikipedia for 680 queries.
Table~\ref{tab:dataset_examples} shows representative examples from each source and class.
\subsection{Data Format}

Each sample in \dataset contains a query, a taxonomy label, and, where applicable, valid interpretations;
each interpretation consists of a clarification, valid answers, and optional supporting context.
Knowledge-dependent ambiguous (\kdamb) samples contain multiple incompatible interpretations,
whereas unambiguous samples (\closed and \open) contain a single interpretation whose answers are mutually compatible.
\closed queries have a finite valid-answer set;
\open queries have an unbounded answer space, so we store one representative valid answer
rather than attempting full enumeration.
Representative answers come from source annotations when available
and are otherwise generated during dataset construction.

Surface-detectable ambiguous (\sdamb) samples do not include answer annotations
because their interpretations arise from an open-ended space of plausible repairs or completions.
For instance, ``\texttt{where is the}'' could be completed as a question about the Colosseum or the Eiffel Tower,
yielding incompatible answers such as Rome or Paris.
Since these cases lack a bounded set of canonical repairs,
we treat \sdamb queries as requiring clarification rather than exhaustively annotating all interpretations and answers.

\subsection{Annotation Protocol and Reliability}

All labels and annotations are produced by the authors,
who are fluent in English.
The authors apply the four-class taxonomy to every query.
Each query from the four sources is manually inspected and verified.
To assess label reliability, two authors independently annotate a shared set of 150 queries
spanning all four classes (50 \sdamb, 35 \kdamb, 35 \closed, and 30 \open).
They achieve 88\% agreement (Cohen's $\kappa = 0.81$).
Remaining disagreements are resolved through discussion or, when necessary, adjudication by a third author.

During annotation, we exclude \kdamb candidates whose interpretations yield identical answer sets.
Although such queries may be entity-ambiguous, they are not ambiguous under our conflict-based definition
when the competing interpretations do not induce incompatible answers.
Since all plausible interpretations support the same answer, a system can answer directly without first asking the user to disambiguate.

% --- ORCAS ---
\subsection{ORCAS Dataset}
\label{app:orcas}
ORCAS~\cite{craswell2020orcas} consists of real-world user search queries.
We use it primarily to source surface-detectable ambiguity,
while also retaining manually verified non-\sdamb cases that satisfy the taxonomy definitions.
Because the full \orcastotal{}-query corpus is large, we use lightweight,
high-recall regex heuristics to retrieve candidate surface defects.
These heuristics target three non-exclusive pattern families, described below.

{\setlength{\leftmargini}{1em}
\setlength{\partopsep}{0pt}
\begin{itemize*}
\pagebreak[4]
\item \textbf{Meaningless noise.}
Queries consisting entirely of non-alphanumeric symbols,
keyboard sequences (e.g., \texttt{asdfghjkl;'\#}),
consonant clusters (e.g., \texttt{fffrff}),
or vowel runs (e.g., \texttt{iuuui})
that render the string unintelligible.
\item \textbf{Structural incompleteness.}
Queries with truncated syntax or missing arguments,
including stub questions (e.g., \texttt{where are}),
dangling determiners or articles (e.g., \texttt{what is the}),
incomplete comparisons (e.g., \texttt{steam vs}),
and fragments ending in a preposition (e.g., \texttt{when to}).
\item \textbf{Referential underspecification.}
Syntactically complete queries
containing pronouns or definite noun phrases
whose referents cannot be resolved from the query alone
(e.g., \texttt{who sings this}, \texttt{what is the movie}, \texttt{can i watch it}).
\end{itemize*}}

\paragraph{Manual verification.}
Heuristic matching on the full dataset yields ${\sim}$\orcascandidates{} candidate queries.
We randomly sample \orcasreview{} candidates and manually label each
as a true surface-detectable case or a false positive;
\orcasconfirmed{} are confirmed as genuine.
Recurring false positives (e.g., \texttt{what kind of word is and}, \texttt{hermes is the god of})
are added to rejection lists and excluded from later sampling.
Candidates filtered out as non-\sdamb are manually relabeled under the full taxonomy rather than discarded.
For \kdamb and \closed cases requiring grounding, we annotate 53 oracle context passages across 46 samples.
For \open cases, whose answer spaces are unbounded, we generate one representative valid answer using the prompt in Table~\ref{tab:prompt_open_answers}.

% --- AmbigNQ ---
\subsection{AmbigNQ Dataset}
AmbigNQ~\cite{min2020ambigqa} provides factoid questions with crowdsourced disambiguations,
as well as single-answer questions from Natural Questions~\cite{kwiatkowski-etal-2019-natural}.
We use the former as candidates for knowledge-dependent ambiguity and the latter as candidates for closed unambiguity.

\paragraph{Conflict-based relabeling criterion.}
AmbigNQ considers a question ambiguous if it admits multiple valid interpretations, each resolved by a disambiguated rewrite.
This definition is well-suited for generating disambiguated question--answer pairs,
but does not match our notion of \emph{logical conflict} among answers.
For example, the query ``Where is the oldest bristlecone pine tree located?''
admits answers at different geographic granularities, such as a peak, state, and country.
We treat these as one interpretation because they identify the same referent at different levels of specificity.
To formalize this distinction, we enforce a conflict-based criterion:
two candidate answers belong to the same interpretation if both can be
included in a single response without making either
incorrect, irrelevant, or misleading.
If including one answer renders the other false or inapplicable,
they constitute distinct, incompatible interpretations.
Annotators were given the following guidelines:

{\setlength{\leftmargini}{1em}
\setlength{\partopsep}{0pt}
\begin{itemize*}
\item \textbf{List merge:} If answers are complementary elements of the same response
(e.g., filming locations for ``Where was Pirates of the Caribbean 2 filmed?''), they form a single interpretation.
\item \textbf{Granularity merge:} If answers identify the same referent at different levels of specificity
(e.g., the peak, state, and country for ``Where is the oldest bristlecone pine tree located?''), they form a single interpretation.
\item \textbf{Aspect merge:} If answers describe facets of the same referent or event
(e.g., the band and vocalist for ``Who sings A Crazy Little Thing Called Love?''), they form a single interpretation.
\end{itemize*}}
%
%--------------------------------------------------------------
\begin{table}[t]
\centering
\small
\caption{\textbf{Summary of AmbigNQ curation.} The top block tracks sample counts
from the initial pool to the retained set; the lower block reports revision operations
on the 1{,}679 retained samples, which may overlap.}
\label{tab:ambignq_curation}
\begin{tabular*}{\columnwidth}{@{\extracolsep{\fill}}lr@{}}
\toprule
Curation & \# Samples \\
\midrule
Initial AmbigNQ samples & 2{,}000 \\
Excluded temporally evolving queries & 321 \\
Retained samples & 1{,}679 \\
\midrule
Added oracle context passages & 196 \\
Corrected wrong or outdated answers & 73 \\
Added missing answers & 56 \\
Relabeled ambiguous $\rightarrow$ unambiguous & 206 \\
Relabeled unambiguous $\rightarrow$ ambiguous & 15 \\
Flagged stable time-dependent queries & 28 \\
Rewrote completed future-tense queries & 13 \\
\bottomrule
\end{tabular*}
\end{table}
%--------------------------------------------------------------
\paragraph{Curation summary.}
Table~\ref{tab:ambignq_curation} summarizes our curation process.
We randomly sample 2{,}000 AmbigNQ queries to enable exhaustive manual inspection
under our conflict-based criterion while maintaining balance across taxonomy classes.
We exclude 321 queries whose answers change over time,
leaving 1{,}679 AmbigNQ-derived samples in \dataset.
Because AmbigNQ's original labels do not always match our conflict-based taxonomy,
we further revise the retained samples.
These revisions fall into two categories:
(1) \textit{data-quality fixes:} we annotate 2{,}409 oracle context passages across 196 factoid samples,
correct outdated or wrong answers, and supply missing answers;
(2) \textit{label corrections:} in addition to 206 ambiguous $\rightarrow$ unambiguous relabelings,
we relabel 15 samples unambiguous $\rightarrow$ ambiguous,
where the original single-answer annotation overlooks a competing interpretation.

\paragraph{Temporal handling.}
The excluded queries are those whose ground truth changes over time.
We nonetheless retain 28 technically time-dependent queries whose answers are expected to
remain relatively stable for the foreseeable future, and flag them for future re-validation
(e.g., ``The most populous country in the world is?''~$\rightarrow$~India).
We also rewrite 13 future-tense queries about completed events
into the past tense, making their answers fixed and reproducible.

%--------------------------------------------------------------
% --- AmbER Generation Prompts ---
\begin{table*}[t]
\centering
\small
\caption{
\textbf{GPT-4o prompts used to instantiate AmbER queries.}
Labels are assigned deterministically from the AmbER entity-group structure:
queries that contain a shared surface name and omit disambiguating descriptors are labeled knowledge-dependent ambiguous.
Queries with disambiguating descriptors are labeled closed unambiguous.
}
\label{tab:amber_prompts}
\begin{tabular}{@{}m{0.20\textwidth}p{0.77\textwidth}@{}}
\toprule
Label & Prompt \\
\midrule
Knowledge-dependent ambiguous
&
\begin{minipage}[c]{0.77\textwidth}
\textbf{Task:} Given an ambiguous entity name and its Wikipedia interpretations,
generate a query that mentions only the shared entity name and could reasonably refer to multiple interpretations.

\textbf{Rules.}
\begin{enumerate}[nosep,leftmargin=1.4em,label=\arabic*.]
\item The query must be genuinely ambiguous across the provided interpretations and must not include disambiguating descriptors.
\item Use the associated properties to choose a plausible query relation that can apply to multiple interpretations.
\item Each answer must be extractable from the corresponding context passage.
\item Return only valid JSON: \{"query": "...", "interpretations": [\{"qid": "...", "wikipedia\_title": "...", "clarification": "...", "answers": ["..."]\}]\}.
\end{enumerate}

\textbf{Input format.} For each ambiguous entity group:\\
Entity name: \texttt{\{name\}}\\
Interpretations, one per entity sharing the surface name:\\
\quad Entity: \texttt{\{title of Wikipedia article\}}\\
\quad Context: \texttt{\{first 3 sentences from Wikipedia article\}}\\
\quad Candidate relations/properties: \texttt{\{property name\} -- \{top-5 values\}}
\end{minipage}
\\
\midrule
Closed unambiguous
&
\begin{minipage}[c]{0.77\textwidth}
\textbf{Task:} Given a target Wikipedia interpretation from an ambiguous entity group,
generate a query that identifies the intended referent.

\textbf{Rules.}
\begin{enumerate}[nosep,leftmargin=1.4em,label=\arabic*.]
\item Add only the minimal descriptor needed to identify the intended referent, such as an occupation, work type, nationality, or domain.
\item Use the associated properties to choose a plausible query relation for the target referent.
\item Each answer must be extractable from the corresponding context passage.
\item Return only valid JSON: \{"query": "...", "label": "closed unambiguous", "answers": ["..."]\}.
\end{enumerate}

\textbf{Input format.} \\
Shared surface name: \texttt{\{name\}}\\
Target referent: \texttt{\{title of Wikipedia article\}}\\
Context: \texttt{\{first 3 sentences from Wikipedia article\}}\\
Associated properties: \texttt{\{property name\} -- \{top-5 values\}}
\end{minipage}
\\
\bottomrule
\end{tabular}
\end{table*}

%--------------------------------------------------------------
% --- AmbER ---
\subsection{AmbER Dataset}
AmbER~\cite{chen-etal-2021-evaluating} contains groups of Wikipedia entities
that share a surface name, such as ``Billy Preston'' for both the basketball player and the musician.
We instantiate queries from these entity groups using GPT-4o~\cite{openai2024gpt4ocard},
following the prompts in Table~\ref{tab:amber_prompts}.
Ambiguous queries mention only the shared underspecified name and remain plausible for multiple entities in the group;
we label them knowledge-dependent ambiguous because resolving the intended referent requires entity-level disambiguation.
Closed unambiguous queries add a minimal descriptor
(e.g., ``the musician'' or ``the basketball player'') to identify the intended referent.
This construction keeps query forms similar across labels while making the distinction depend on entity-level disambiguation
rather than template-level artifacts.
We manually verify every query for label correctness and entity plausibility,
annotating 786 oracle context passages across 400 samples by checking that each passage supports the intended entity.

%--------------------------------------------------------------
% --- Dolly-15K ---
\subsection{Dolly-15K Dataset}
Dolly-15K~\cite{DatabricksBlog2023DollyV2} contains human-generated
instruction--response pairs across diverse task categories.
We specifically filter for general QA and brainstorming tasks,
which primarily yield open unambiguous queries with multiple compatible responses.
Queries with a closed answer space, such as a fixed list or single factual answer, are labeled closed unambiguous.
The authors manually verify this closed-versus-open assignment for every query.
We additionally annotate 51 oracle context passages across 38 queries.
%

% Ambiguation prompt for AmbigTriviaQA dataset construction
\begin{table}[t]
\centering
\small
\caption{GPT-4o ambiguation prompt for constructing the \mbox{AmbigTriviaQA} dataset.}
\begin{tabular}{p{0.9\columnwidth}}
\toprule
Please make the following question ambiguous. \\
Your task is to introduce ambiguity by altering the specificity of the noun phrase
or omitting crucial details from the statement.
Keep the rest of the sentence unchanged except for the modified sections.\\
Generate only the revised statement. \\
Question: \texttt{\{question\}} \\
Ambiguation: \\
\bottomrule
\end{tabular}
\label{tab:prompt_ambiguate}
\end{table}

% Validation prompt for generated AmbigTriviaQA queries
\begin{table}[t]
\centering
\small
\caption{GPT-4o verification prompt for constructing the \mbox{AmbigTriviaQA} dataset.}
\begin{tabular}{p{0.9\columnwidth}}
\toprule
An ambiguous question has multiple valid answers. \\
Is the following question ambiguous with multiple possible answers?
Answer only in Yes or No. \\
Question: \texttt{\{ambiguous generation\}} \\
Yes or No: \\
\bottomrule
\end{tabular}
\label{tab:prompt_verify}
\end{table}

% Example ambiguation pairs in AmbigTriviaQA
\begin{table*}[t]
\centering
\small
\caption{
\textbf{Example ambiguation pairs in \mbox{AmbigTriviaQA}.}
The ambiguated phrase is highlighted in bold.}
\label{tab:ambigtrivia_examples}
\begin{tabularx}{\textwidth}{@{}>{\centering\arraybackslash}X >{\centering\arraybackslash}X@{}}
\toprule
Original Question & Ambiguated Question \\
\midrule
What is \textbf{Uma Thurman's} middle name? &
What is \textbf{the} middle name? \\
\addlinespace
Who directed \textbf{Back to the Future}? &
Who directed \textbf{that movie}? \\
\addlinespace
In which year was \textbf{CNN} founded? &
When was \textbf{it} founded? \\
\bottomrule
\end{tabularx}
\end{table*}
%--------------------------------------------------------------
% AmbigTriviaQA Construction
%--------------------------------------------------------------
\section{AmbigTriviaQA Construction}
\label{app:dataset_construction}
We build AmbigTriviaQA as an out-of-distribution evaluation set by deriving
ambiguous questions from TriviaQA~\cite{DBLP:journals/corr/JoshiCWZ17}, adapting the
procedure of~\citet{kim2024apa}.
Each original question is first rewritten into an ambiguous form by GPT-4o using
the template in Table~\ref{tab:prompt_ambiguate}.
GPT-4o then re-examines every rewrite (Table~\ref{tab:prompt_verify}) as an automatic
quality filter, and any rewrite it does not judge ambiguous is discarded.
Example ambiguation pairs are shown in Table~\ref{tab:ambigtrivia_examples}.
To mitigate confirmation bias from using the same model for generation and filtering,
the authors review the surviving questions and curate the final set.
The resulting set contains 1{,}000 queries, evenly split into 500 ambiguous and 500 unambiguous.
%

%--------------------------------------------------------------
% Implementation Details
%--------------------------------------------------------------
\section{Implementation Details}
\label{app:impl_details}
We split \dataset into train/validation/test sets (3{,}762/470/471; 8:1:1).
The conflict modeling module is trained on the non-\sdamb subset ($\sim$2{,}540 samples per seed).
At evaluation, the test split is routed end-to-end through the early-exit and conflict modeling modules.
Among the test queries, 218 have oracle context passages; we call these the factoid subset.
The \emph{factoid} setting uses this 218-query subset:
Table~\ref{tab:binary} follows prior work with context-free candidate-answer generation,
while Table~\ref{tab:context} compares no context, DPR-retrieved context~\cite{karpukhin2020densepassageretrievalopendomain}, and oracle context.
The \emph{all-query} setting (Tables~\ref{tab:4way}, \ref{tab:efficiency}, and~\ref{tab:ablation})
covers the full test split and grounds candidate-answer generation in context when available.

The early-exit module trains a linear head on a frozen DeBERTa-v3-base encoder
for 8 epochs using AdamW~\cite{loshchilov2019decoupledweightdecayregularization}
with learning rate $2{\times}10^{-4}$, weight decay $0.01$, and gradient clipping 1.0.
The conflict modeling module keeps the encoder frozen and trains only lightweight heads
for 200 epochs using AdamW (learning rate $10^{-4}$, weight decay $0.01$).
We use batch size 16 for both modules, retain $k{=}10$ answers after deduplication,
and set $\lambda_{\text{inv}} {=} 1.0$.
Pairwise relations are computed using a frozen DeBERTa-v3-base NLI model fine-tuned on
MNLI~\cite{williams2018broadcoveragechallengecorpussentence}, FEVER~\cite{thorne2018feverlargescaledatasetfact},
and ANLI~\cite{nie2020adversarialnlinewbenchmark}.
Candidate-answer generation uses five LLM calls at temperature 0.7.
When fewer than $k$ unique answers are produced, we fill remaining slots by frequency-proportional duplication;
if parsing fails, all slots are set to \texttt{Unknown} (< 0.5\% of queries).
Distractor generation replaces one duplicate slot with an LLM-generated distractor sampled at temperature 0.8.
We implement models in PyTorch~\cite{paszke2019pytorchimperativestylehighperformance} 2.10
with Hugging Face Transformers~\cite{wolf-etal-2020-transformers} 4.57.6
and vLLM~\cite{kwon2023efficientmemorymanagementlarge} 0.19.1.
We run experiments on one node with 2$\times$ NVIDIA B200 GPUs.
%
%--------------------------------------------------------------
\begin{figure}[t]
\centering
\includegraphics[width=\columnwidth]{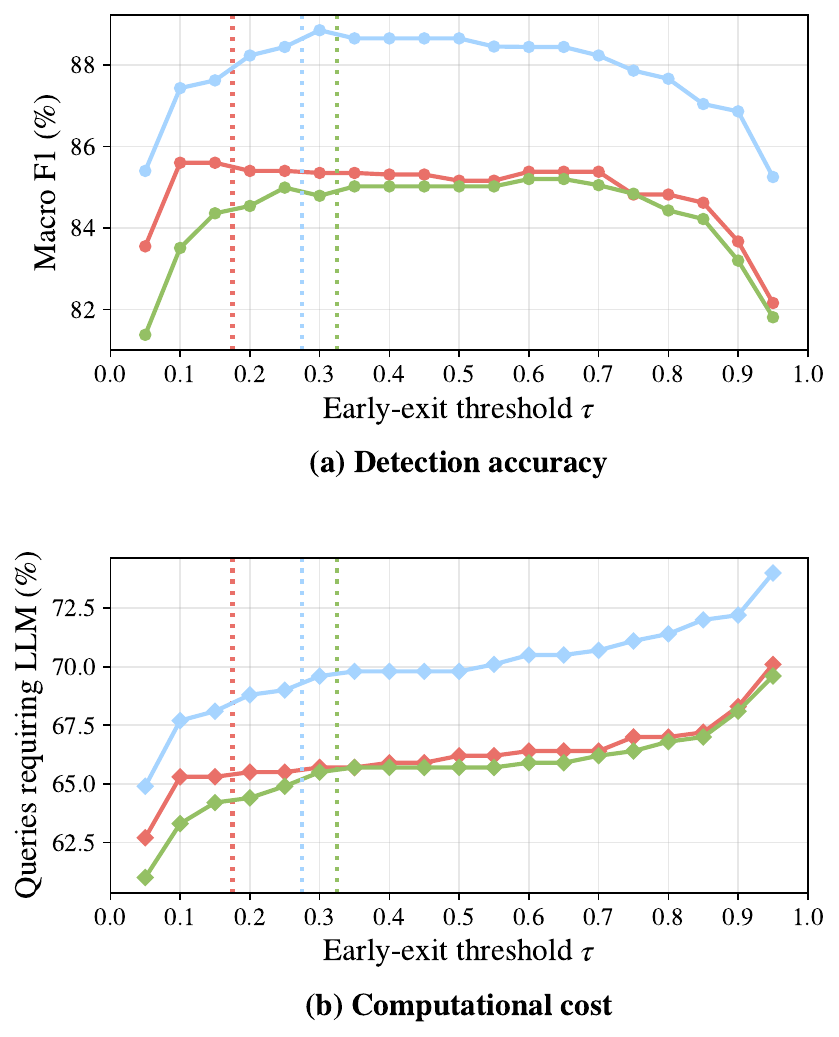}
\caption{
\textbf{\method is robust to early-exit threshold selection.}
Macro-F1 remains stable across a broad range of early-exit thresholds (a),
while higher thresholds route more queries to the LLM (b).
Colored curves show per-seed sweeps using Qwen2.5-14B-Instruct;
dotted vertical lines indicate validation-selected thresholds;
and cropped y-axes highlight threshold-dependent trends.
}
\label{fig:threshold_sweep}
\end{figure}
%--------------------------------------------------------------
% Hyperparameter Sensitivity
%--------------------------------------------------------------

% Answer generation prompts
\begin{table*}[t]
\centering
\small
\caption{\textbf{Prompts for candidate answer generation.} The context-grounded prompt is used when oracle context is available;
otherwise, the context-free prompt is used.
Three examples for each prompt are omitted for brevity.}
\label{tab:generation_prompts}
\begin{tabular}{@{}m{0.19\textwidth}p{0.78\textwidth}@{}}
\toprule
Setting & Prompt \\
\midrule
Context-grounded
&
\begin{minipage}[c]{0.78\textwidth}
\textbf{Task:} You will receive a question and grounding context.
Produce a JSON list of candidate answers.

\textbf{Rules.}
\begin{enumerate}[nosep,leftmargin=1.4em,label=\arabic*.]
\item Identify all plausible candidate answers supported by the context.
\item If the context supports multiple dates, locations, entities, or facets, extract each as a separate candidate answer.
\item Each answer must be a short, independent, declarative sentence.
\item Return only valid JSON: \{"answers": ["...", "...", ...]\}.
\item The answers list must contain at least one string.
\item Do not include meta-commentary such as ``The context does not provide...'' or ``I cannot determine...''.
\end{enumerate}
Question: \texttt{\{question\}} \\
Context: \texttt{\{context\}} \\
Answers:
\end{minipage}
\\
\midrule
Context-free
&
\begin{minipage}[c]{0.78\textwidth}
\textbf{Task:} You will receive a question.
Produce a JSON list of candidate answers.

\textbf{Rules.}
\begin{enumerate}[nosep,leftmargin=1.4em,label=\arabic*.]
\item Identify all plausible candidate answers using world knowledge.
\item For list-type queries, give each item as a separate candidate answer.
\item Each answer must be a short, independent, declarative sentence.
\item Return only valid JSON: \{"answers": ["...", "...", ...]\}.
\item The answers list must contain at least one string.
\item Do not include meta-commentary such as ``This is subjective...'' or ``I cannot determine...''.
\end{enumerate}
Question: \texttt{\{question\}} \\
Answers:
\end{minipage}
\\
\bottomrule
\end{tabular}
\end{table*}

%--------------------------------------------------------------
\section{Hyperparameter Sensitivity}
\label{app:hyperparameter}

\subsection{Early-exit Threshold}
Figure~\ref{fig:threshold_sweep} reports per-seed test-set sweeps of the early-exit routing threshold $\tau$ over $[0.05, 0.95]$
using Qwen2.5-14B-Instruct checkpoints, with the full \method pipeline evaluated end-to-end.
The threshold trades off early-exit efficiency and detection accuracy;
the operating value is selected on the validation split to maximize the early-exit module's binary \sdamb-detection F1.
Panel~(a) shows that 4-way macro-F1 remains stable across a broad middle range of $\tau$ and degrades at the extremes.
At low $\tau$, the early-exit module routes too aggressively and
many non-\sdamb queries are incorrectly early-exited as \sdamb.
At high $\tau$, genuine \sdamb queries bypass early exit and reach the conflict modeling module,
which has no \sdamb class and misclassifies them.
All three seeds exhibit the same qualitative pattern, indicating that the plateau and extreme-case degradation reflect the routing mechanism rather than a seed-specific artifact.
Panel~(b) shows that the fraction of queries requiring LLM inference rises with $\tau$ in every seed.
The validation-selected thresholds (mean $\tau \approx 0.26$, range $0.175$--$0.325$)
fall within each seed's accuracy plateau, achieving mean macro-F1 of $85.37 \pm 2.42$\%.
This indicates that validation tuning selects thresholds aligned with downstream performance.
%--------------------------------------------------------------
% Distractor generation prompt
\begin{table*}[t]
\centering
\small
\caption{\textbf{Prompt for generating distractors.} The \texttt{Context:} field is included only when oracle context is available.}
\label{tab:prompt_distractor}
\begin{tabular}{@{}p{\textwidth}@{}}
\toprule
\makebox[\textwidth][c]{%
\begin{minipage}{0.95\textwidth}
\textbf{Task.}
Generate three distractor answers for the given question.
Each should be plausible, but verifiably invalid with respect to the provided valid answers and, when available, context.

\textbf{Allowed distractor types.}
\begin{enumerate}[nosep,leftmargin=1.4em,label=\arabic*.]
\item Partial answer: Omit a required component from a multi-part valid answer.
\item Role swap or misattribution: Reassign entities from the context to incorrect roles or relationships.
\item Controlled perturbation: Modify a date, number, name, or attribute appearing in the context or valid answers.
\end{enumerate}

\textbf{Constraints.}
\begin{itemize}[nosep,leftmargin=1.4em,label=\textbullet]
\item Do not introduce real-world entities, items, or facts absent from the provided information.
\item Do not generate paraphrases, close synonyms, or valid subsets of the valid answers.
\item Each distractor must be grounded in the provided information but altered so that it is invalid.
\item Match the style, tone, and sentence structure of the valid answers.
\item Use at least two distinct distractor types across the three outputs.
\item Return only valid JSON: \{"distractors": ["...", "...", "..."]\}.
\end{itemize}
Question: \texttt{\{question\}} \\
Context: \texttt{\{context, if available\}} \\
Generated Valid Answers: \texttt{\{valid answers\}} \\
Distractors:
\end{minipage}}
\\
\bottomrule
\end{tabular}
\end{table*}

% Example distractors
\begin{table*}[t]
\centering
\small
\setlength{\tabcolsep}{4pt}
\renewcommand{\arraystretch}{1.15}
\caption{
\textbf{Example distractors.}
One representative distractor per query; the bold span marks the perturbation
that makes it invalid.}
\label{tab:distractor_examples}
\renewcommand{\tabularxcolumn}[1]{m{#1}}
\begin{tabularx}{\textwidth}{
@{}>{\RaggedRight\arraybackslash}m{0.22\textwidth}
   >{\RaggedRight\arraybackslash}m{0.34\textwidth}
   >{\RaggedRight\arraybackslash}X@{}}
\toprule
Query & Valid Answer(s) & Distractor \\
\midrule

\multicolumn{3}{@{}l}{\textit{Knowledge-dependent ambiguous}} \\
\cmidrule(lr){1-3}
\makecell[l]{When did the Simpsons\\first air on television?} &
\makecell[l]{April 19, 1987 (animated short),\\
December 17, 1989 (prime time show)} &
\makecell[l]{The Simpsons debuted as a half-hour show on\\
December 17, \textbf{1990}.} \\

\addlinespace
\midrule
\multicolumn{3}{@{}l}{\textit{Closed unambiguous}} \\
\cmidrule(lr){1-3}
\makecell[l]{Who directed the film\\New York, New York?} &
Martin Scorsese &
\makecell[l]{\textbf{Earl Mac Rauch} directed the film New York,\\
New York.} \\

\addlinespace
\midrule
\multicolumn{3}{@{}l}{\textit{Open unambiguous}} \\
\cmidrule(lr){1-3}
\makecell[l]{How can I sleep well\\at night?} &
\makecell[l]{Maintain a consistent sleep schedule\\
Avoid caffeine and large meals before bed} &
\makecell[l]{You can sleep well at night by going to bed and\\
waking up \textbf{at different times every day.}} \\

\bottomrule
\end{tabularx}
\end{table*}

\begin{figure}[t]
\centering
\includegraphics[width=\columnwidth]{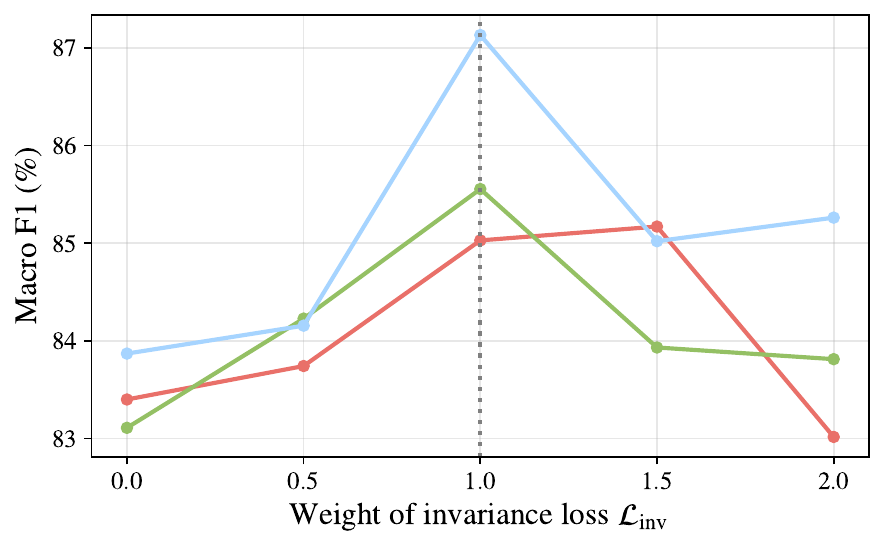}
\caption{
\textbf{\method benefits from the invariance loss, with $\lambda_{\mathrm{inv}}{=}1.0$ providing a stable high-performing choice.}
Colored curves show per-seed 4-way macro-F1 on the \dataset test split using Qwen2.5-14B-Instruct;
the dotted vertical line marks the selected value.
}
\label{fig:lambda_inv_sweep}
\end{figure}
%--------------------------------------------------------------
\subsection{Strength of Invariance Loss}
We sweep the weight of the invariance loss
$\lambda_{\mathrm{inv}} \in \{0, 0.5, 1.0, 1.5, 2.0\}$ on the \dataset test split
using the same three Qwen2.5-14B-Instruct seeds as in the main experiments.
Figure~\ref{fig:lambda_inv_sweep} reports per-seed 4-way macro-F1.
We make two observations.
First, the invariance loss improves performance:
$\lambda_{\mathrm{inv}}{=}1.0$ outperforms $\lambda_{\mathrm{inv}}{=}0$ for all three seeds
and yields the best cross-seed mean in the sweep.
This trend is consistent with the ``w/o~$\mathcal{L}_{\text{inv}}$'' row of Table~\ref{tab:ablation}.
Second, \method does not require fine-grained tuning of $\lambda_{\mathrm{inv}}$: nearby nonzero values stay close to the best setting,
although large weights degrade some seeds.
We select $\lambda_{\mathrm{inv}}{=}1.0$, which lies in a high-performing region,
suggesting that the invariance objective provides robust regularization.
%--------------------------------------------------------------
% Generation Prompts
%--------------------------------------------------------------
\newpage
\section{Generation Prompts}
\label{app:prompts}
Our pipeline uses three system-level LLM prompts:
two answer-generation prompts (Table~\ref{tab:generation_prompts}), selected by the availability of context,
and one distractor-generation prompt (Table~\ref{tab:prompt_distractor}).
Both answer-generation prompts include three examples.
In the grounded setting, the user message supplies the query and passage;
the prompt asks the LLM to extract all answers supported by that passage.
In the context-free setting, the LLM is instructed to generate candidate answers from world knowledge.

We generate distractors using the same frozen LLM and the prompt in Table~\ref{tab:prompt_distractor}.
Given the query, passage, and valid answers produced by the candidate-answer generation prompt,
the LLM generates distractors that match the surface form, style, and format of valid answers
but are factually invalid given the query and passage.
For each training example, we randomly select one of the three generated distractors
to replace a duplicate answer slot, yielding the augmented answer set.
In \dataset, candidate generation always yields fewer than $k{=}10$ unique answers.
Consequently, every answer set contains at least one duplicate slot.
Distractors apply only to the three routed classes;
surface-detectable ambiguous queries exit early.
We show examples of distractors in Table~\ref{tab:distractor_examples}.

%--------------------------------------------------------------
% Taxonomy Utility Experiment
%--------------------------------------------------------------
\section{Taxonomy Utility Experiment}
\label{app:taxonomy_exp}

% The four taxonomy definition blocks
\begin{table*}[t]
\centering
\small
\caption{
\textbf{Taxonomy definitions used in the utility experiment.}
The four \texttt{Taxonomy:} blocks are substituted into the scaffold of Table~\ref{tab:prompt_taxonomy_scaffold}.
The (A)/(U) tags shown here mark the ambiguous/unambiguous mappings, respectively, used for binary F1;
they are added for clarity and are not part of the prompt seen by the model.
}
\label{tab:taxonomy_defs}
\begin{tabularx}{\textwidth}{@{}l X@{}}
\toprule
Taxonomy & Class Definitions \\
\midrule
\makecell[tl]{\textbf{\method}\\\textbf{(proposed)}} &
\textbf{(A) Surface-detectable ambiguity:} The query is ambiguous, incomplete, malformed, or underspecified in a way that is detectable from the query text alone, without external knowledge. \newline
\textbf{(A) Knowledge-dependent ambiguity:} The query is well-formed, but recognizing its multiple legitimate interpretations requires external or world knowledge. \newline
\textbf{(U) Closed unambiguity:} The query is clear and has a single definitive answer. No clarification is needed. \newline
\textbf{(U) Open unambiguity:} The query is clear and allows diverse but compatible valid answers, such as advice, examples, or lists. No clarification is needed. \\
\addlinespace

\makecell[tl]{Symptom-based\\\citep{zhang2024clamber}} &
\textbf{(A) Unfamiliar:} The query contains unfamiliar entities or facts. \newline
\textbf{(A) Contradiction:} The query contains self-contradictions. \newline
\textbf{(A) Lexical:} The query contains terms with multiple meanings. \newline
\textbf{(A) Semantic:} The query lacks context, leading to multiple interpretations. \newline
\textbf{(A) Who:} The query is underspecified with respect to a person, entity, or agent needed to answer it. \newline
\textbf{(A) When:} The query is underspecified with respect to time or temporal scope. \newline
\textbf{(A) Where:} The query is underspecified with respect to place, location, or spatial scope. \newline
\textbf{(A) What:} The query is underspecified with respect to the requested information, object, or task. \newline
\textbf{(U) Unambiguous:} The query is clear, specific, and unambiguous. \\
\addlinespace

\makecell[tl]{Action-oriented\\\citep{tang2025clarifying}} &
\textbf{(A) Semantic:} The query is semantically ambiguous for several common reasons: it may include homonyms;
a word in the query may refer to a specific entity while also functioning as a common word;
or an entity mentioned in the query could refer to multiple distinct entities. \newline
\textbf{(A) Generalize:} The query focuses on specific information;
however, a broader, closely related query might better capture the user's true information needs. \newline
\textbf{(A) Specify:} The query has a clear focus but may encompass too broad a research scope.
It is possible to further narrow down this scope by providing more specific information related to the query. \newline
\textbf{(U) Unambiguous:} The query is clear, specific, and unambiguous. \\
\addlinespace

\makecell[tl]{Linguistic\\\citep{tanjim2025detecting}} &
\textbf{(A) Pragmatic:} The meaning of the query depends on missing context, reference, or scope. \newline
\textbf{(A) Syntactic:} The structure of the query is incomplete or allows multiple interpretations. \newline
\textbf{(A) Lexical:} The meaning of a word or term in the query is unclear or has multiple interpretations. \newline
\textbf{(U) Unambiguous:} The query is clear, specific, and unambiguous. \\
\bottomrule
\end{tabularx}
\end{table*}
%--------------------------------------------------------------
In Section~\ref{sec:exp_taxonomy}, we evaluate whether the taxonomy improves
ambiguity classification and response quality over existing taxonomies.
We compare \method against three alternatives:
symptom-based~\citep{zhang2024clamber},
action-oriented~\citep{tang2025clarifying},
and linguistic~\citep{tanjim2025detecting}.
We exclude \citet{min2020ambigqa}, \citet{guo2021abgcoqa}, and
\citet{amplayo2022queryrefinementpromptsclosedbook}, whose taxonomies target
dataset-specific subsets rather than general prompt-based ambiguity detection.
We use a controlled prompt design to isolate the effect of the taxonomy.
Specifically, we embed per-taxonomy class descriptions from each paper (Table~\ref{tab:taxonomy_defs})
in a shared prompt scaffold (Table~\ref{tab:prompt_taxonomy_scaffold}),
varying only the \texttt{Taxonomy:} block.
For \citet{zhang2024clamber}'s Who/When/Where/What classes,
we rephrase the original template (``Query output contains confusion due to missing $\ldots$ elements'')
as ``The query is underspecified with respect to $\ldots$'' for clearer LLM prompting.
Because the baseline taxonomies do not define an unambiguous class, we use a common description for it across all schemas.

We evaluate each taxonomy along two axes: binary ambiguity detection and response quality.
For each taxonomy, Qwen2.5-14B-Instruct receives oracle context
and generates one output per query using greedy decoding with a 150-token cap.
As shown in Table~\ref{tab:prompt_taxonomy_scaffold},
each output contains a classification and a response:
a direct answer for queries predicted as unambiguous,
or a clarifying question for those predicted as ambiguous.
%
%--------------------------------------------------------------
% Shared scaffold for the taxonomy utility experiment
\begin{table}[t]
\centering
\small
\caption{
\textbf{Shared prompt scaffold for the taxonomy utility experiment.}
The \texttt{Taxonomy:} placeholder is replaced with taxonomy class definitions from Table~\ref{tab:taxonomy_defs}.
}
\label{tab:prompt_taxonomy_scaffold}
\begin{tabular}{@{}p{0.95\columnwidth}@{}}
\toprule
Classify the query using the provided taxonomy, then produce the corresponding response. \\
If the query is unambiguous, answer it directly.\\
If the query is ambiguous, ask a clarifying question that would help resolve the ambiguity.

You must format your output exactly as:\\
\hspace*{2em}Classification: <class>\\
\hspace*{2em}Response: <answer or clarifying question>

Taxonomy: \texttt{\{taxonomy-specific class definitions\}}

Query: \texttt{\{query\}}\\
Context: \texttt{\{context, if available\}}

Classification:\\
Response:\\
\bottomrule
\end{tabular}
\vspace{-1em}
\end{table}
%--------------------------------------------------------------
%
For ambiguity detection, we map each taxonomy class to a binary label and compute F1.
For response quality, we evaluate classes with well-defined references:
closed unambiguous queries are scored with Ans BERTScore against gold answers,
whereas knowledge-dependent ambiguous queries use CQ BERTScore against pseudo-references.
Ans BERTScore measures whether the model's answer is semantically equivalent to the gold answer, while
CQ BERTScore measures whether the model's clarifying question matches a question that helps the user choose among concrete interpretations.
We generate pseudo-references by applying the clarification prompt of \citet{kim2024apa}
in Table~\ref{tab:prompt_cq_ref} to each query and its gold disambiguation.
Conditioning on the gold disambiguation ensures the pseudo-reference
asks about the ambiguity a valid clarifying question should resolve.
%--------------------------------------------------------------
% CQ pseudo-reference generation prompt
\begin{table}[t]
\centering
\small
\caption{
\textbf{Prompt for generating pseudo-reference clarifying questions.}
We adapt the prompt of \citet{kim2024apa} to generate a pseudo-reference
from each ambiguous query and its gold disambiguation.
}
\label{tab:prompt_cq_ref}
\begin{tabular}{@{}p{0.96\columnwidth}@{}}
\toprule
Engage with the provided ambiguous question by extracting the key point of ambiguity,
and interactively ask for clarification based on the disambiguated question.

\vspace{4pt}
Example 1: \\
Ambiguous Question: Who won? \\
Disambiguation: Who won the 2020 U.S. presidential election? \\
Clarification Request: Your question seems ambiguous. Could you specify which competition or event you are asking about?

\vspace{4pt}
Example 2: \\
Ambiguous Question: What's the weather like? \\
Disambiguation: What's the weather like in Miami today? \\
Clarification Request: Your question is ambiguous. Where are you interested in the weather report for?

\vspace{4pt}
Ambiguous Question: \texttt{\{query\}} \\
Disambiguation: \texttt{\{gold disambiguation\}} \\
Clarification Request:
\\
\bottomrule
\end{tabular}
\end{table}

% Example clarifying question pseudo-references
\begin{table}[!t]
\centering
\small
\caption{
\textbf{Examples of pseudo-references for clarifying questions.}
Knowledge-dependent ambiguous (KDAmb) references target concrete disambiguations,
whereas surface-detectable ambiguous (\sdamb) references reduce to generic requests for specification.
}
\label{tab:cq_examples}
\renewcommand{\arraystretch}{1.15}
\renewcommand{\tabularxcolumn}[1]{m{#1}}
\begin{tabularx}{\columnwidth}{
@{}>{\centering\arraybackslash}m{0.13\columnwidth}
   >{\RaggedRight\arraybackslash}m{0.29\columnwidth}
   >{\RaggedRight\arraybackslash}X@{}}
\toprule
\textbf{Class} & \textbf{Query} & \textbf{Generated reference CQ} \\
\midrule

\multirow[c]{2}{=}{\raisebox{-2.3em}{\centering\arraybackslash KDAmb}}
& When did the Simpsons first air on television?
& Could you please specify whether you mean its debut as an animated short or its first airing as a half-hour prime-time show? \\
\cmidrule(l){2-3}
& What does Billy Preston play?
& Could you specify whether you mean the singer-songwriter or the professional basketball player? \\

\midrule

\multirow[c]{2}{0.13\columnwidth}{\raisebox{-2.3em}{\centering \sdamb}}
& iuuui
& Could you provide more context or specify what you would like to know? \\
\cmidrule(l){2-3}
& what is the
& Could you please specify what exactly you would like to know or learn about? \\

\bottomrule
\end{tabularx}
\end{table}
%--------------------------------------------------------------
%
Open unambiguous and surface-detectable ambiguous queries are excluded from response-quality scoring because they lack informative finite references.
The unbounded answer space of open unambiguous queries admits no finite gold set.
The references for surface-detectable ambiguous queries collapse to generic requests for more context or specification,
rather than targeted clarifying questions that ask the user to choose among concrete interpretations (see Table~\ref{tab:cq_examples} for examples).
%
%--------------------------------------------------------------
% Competitor Details
%--------------------------------------------------------------
% Direct baseline prompt
\begin{table}[t]
\centering
\small
\caption{Direct baseline prompt.}
\label{tab:prompt_direct}
\begin{tabular}{@{}p{0.95\columnwidth}@{}}
\toprule
Answer the following question.\\
Question: \texttt{\{question\}}\\
Context: \texttt{\{context, if available\}}\\
Answer:\\
\bottomrule
\end{tabular}
\end{table}

% Ambig-Aware baseline prompt
\begin{table}[t]
\centering
\small
\caption{Ambig-Aware baseline prompt.}
\label{tab:prompt_ambig_aware}
\begin{tabular}{@{}p{0.95\columnwidth}@{}}
\toprule
Answer the following question.\\
If the question is ambiguous, it is proper to answer with "The question is ambiguous".\\
Question: \texttt{\{question\}}\\
Context: \texttt{\{context, if available\}}\\
Answer:\\
\bottomrule
\end{tabular}
\end{table}

% Self-Ask baseline prompt (Step 2)
\begin{table}[!t]
\centering
\small
\caption{Self-Ask step 2 prompt.}
\label{tab:prompt_self_ask}
\begin{tabular}{@{}p{0.95\columnwidth}@{}}
\toprule
Given the question and answer, is the question ambiguous or unambiguous?\\
Answer only ambiguous or unambiguous.\\
Question: \texttt{\{question\}}\\
Answer: \texttt{\{generated\_answer\}}\\
Ambiguous or Unambiguous:\\
\bottomrule
\end{tabular}
\end{table}

% CoT w/ taxonomy baseline prompt
\begin{table*}[!t]
\centering
\small
\caption{\textbf{CoT w/ taxonomy baseline prompt.} Examples are omitted for brevity.}
\label{tab:prompt_cot_taxonomy}
\begin{tabular}{@{}p{0.95\textwidth}@{}}
\toprule
Classify the following query using the ARCHIVE taxonomy.
First reason step by step about whether the query is ambiguous and, if so,
whether the ambiguity is surface-detectable or knowledge-dependent.
Then output the final classification.
\medskip

\textbf{Taxonomy}
\begin{enumerate}[nosep,leftmargin=*,label=\arabic*.]
\item Surface-detectable ambiguity:
The query is ambiguous, incomplete, malformed, or underspecified in a way that is detectable from the query text alone, without external knowledge.
\item Knowledge-dependent ambiguity:
The query is well-formed, but recognizing its multiple legitimate interpretations requires external or world knowledge.
\item Closed unambiguity:
The query is clear and has a single definitive answer. No clarification is needed.
\item Open unambiguity:
The query is clear and allows diverse but compatible valid answers, such as advice, examples, or lists. No clarification is needed.
\end{enumerate}

\medskip
\textbf{Task}
Query: \texttt{\{query\}}\newline
Context: \texttt{\{context, if available\}}\newline
Rationale:\\
Classification:\\
\bottomrule
\end{tabular}
\end{table*}
%--------------------------------------------------------------
\section{Competitor Details}
\label{app:competitor_details}
We provide detailed descriptions of each baseline.
For all methods, oracle context is inserted after the question when available.

\paragraph{Direct.}
Following~\citet{kim2024apa}, the LLM receives a plain QA prompt.
A query is detected as ambiguous if the model's greedy generation contains any of the
predefined ambiguity-indicating phrases---``ambiguous'', ``ambig'', ``unclear'',
``not clear'', ``not sure'', ``confused'', ``confusing'', ``vague'', ``uncertain'',
``doubtful'', ``doubt'', ``questionable'', and ``clarify''---and unambiguous otherwise.
The prompt is shown in Table~\ref{tab:prompt_direct}.

\paragraph{Ambig-Aware~\citep{kim2024apa}.}
The prompt, as shown in Table~\ref{tab:prompt_ambig_aware}, explicitly describes how to handle ambiguity.

\paragraph{Self-Ask.}
A two-step approach detects ambiguity after answer generation.
In step~1, the LLM generates a free-form answer to the query using the Direct baseline prompt
(Table~\ref{tab:prompt_direct}).
In step~2, the LLM receives the query and its own answer,
and judges whether the query is ambiguous (Table~\ref{tab:prompt_self_ask}).
Following \citet{kim2024apa}, we modify the prompt from \citet{amayuelas2024knowledge}
so that the model focuses specifically on ambiguity.

\paragraph{CoT w/ taxonomy.}
The baseline prompts the LLM with \method's four-class taxonomy
and four in-context examples, one per class.
As Table~\ref{tab:prompt_cot_taxonomy} shows, the LLM outputs
a step-by-step justification followed by a classification label.
%--------------------------------------------------------------

% Frequency and Entropy Heuristic prompt
\begin{table}[!t]
\centering
\small
\caption{Frequency and entropy heuristics prompt.}
\label{tab:prompt_heuristic}
\begin{tabular}{@{}p{0.95\columnwidth}@{}}
\toprule
Provide all plausible responses to the question.\\
Question: \texttt{\{question\}}\\
Context: \texttt{\{context, if available\}}\\
Answer(s):\\
\bottomrule
\end{tabular}
\end{table}

% Context quality sensitivity table
\begin{table*}[!h]
\centering
\small
\caption{
\textbf{Context improves relation-based ambiguity detection.}
Each cell reports \Famb or \Funamb on factoid queries.
The no-context setting is omitted for \citet{shi2025trustnlp} as their method assumes context is available.
}
\label{tab:context}
\begin{tabular*}{\textwidth}{@{}l@{\extracolsep{\fill}}
C{0.085\textwidth} C{0.085\textwidth}
C{0.085\textwidth} C{0.085\textwidth}
C{0.085\textwidth} C{0.085\textwidth}@{}}
\toprule
\multirow{2}{*}{Method}
 & \multicolumn{2}{c}{No Context}
 & \multicolumn{2}{c}{DPR top-3}
 & \multicolumn{2}{c}{Oracle Context} \\
\cmidrule(lr){2-3}\cmidrule(lr){4-5}\cmidrule(lr){6-7}
 & \Famb & \Funamb & \Famb & \Funamb & \Famb & \Funamb \\
\midrule
\multicolumn{7}{@{}l}{Backbone: \textit{LLaMA-2-7B}} \\
\midrule[0.3pt]
Direct & 2.77 & 59.52 & 2.66 & 58.39 & 6.23 & \underline{79.99} \\
Ambig-Aware & 49.70 & 0.00 & 52.70 & 0.00 & 55.15 & 3.04 \\
Self-Ask & 33.98 & 64.04 & 33.90 & 61.75 & 53.68 & 68.89 \\
CoT w/ taxonomy & 49.85 & 4.01 & 54.44 & 12.94 & \underline{64.62} & 24.49 \\
Frequency & 49.70 & 0.00 & 54.92 & 0.00 & 53.70 & 0.00 \\
Entropy & 48.80 & 32.14 & 49.34 & 1.66 & 55.05 & 34.19 \\
Intent-Sim & 41.84 & \underline{66.57} & 0.29 & \underline{68.32} & 45.35 & \underline{79.99} \\
Shi et al. & \rule[0.5ex]{1em}{0.4pt} & \rule[0.5ex]{1em}{0.4pt} & 43.25 & 45.29 & 53.56 & 54.09 \\
APA & \underline{52.24} & 20.94 & \underline{56.14} & 63.48 & 59.45 & 60.52 \\
\midrule
\textbf{\method (proposed)} & \textbf{58.96} & \textbf{85.57} & \textbf{62.33} & \textbf{77.48} & \textbf{68.31} & \textbf{87.62} \\
\midrule
\multicolumn{7}{@{}l}{Backbone: \textit{Qwen2.5-14B}} \\
\midrule[0.3pt]
Direct & 3.74 & 68.76 & 27.49 & 56.80 & 53.82 & 82.23 \\
Ambig-Aware & 45.32 & 69.18 & 48.35 & 76.22 & 66.94 & 72.08 \\
Self-Ask & 34.51 & \underline{72.43} & 35.39 & 72.59 & \underline{74.31} & 75.75 \\
CoT w/ taxonomy & 45.27 & 69.87 & \underline{62.32} & \textbf{83.02} & 60.55 & \underline{83.79} \\
Frequency & 49.70 & 0.00 & 54.29 & 0.00 & 65.02 & 0.28  \\
Entropy & 49.67 & 2.42 & 49.96 & 17.39 & 64.92 & 7.49 \\
Intent-Sim & 48.68 & 15.91 & 49.60 & 39.86 & 66.22 & 43.83 \\
Shi et al. & \rule[0.5ex]{1em}{0.4pt} & \rule[0.5ex]{1em}{0.4pt} & 43.57 & 39.51 & 50.60 & 51.18 \\
APA & \underline{50.88} & 64.05 & 44.41 & 70.15 & 64.05 & 72.87 \\
\midrule
\textbf{\method (proposed)} & \textbf{61.31} & \textbf{85.66} & \textbf{76.83} & \underline{80.51} & \textbf{76.84} & \textbf{88.35} \\
\bottomrule
\vspace{4pt}
\end{tabular*}
\end{table*}
%--------------------------------------------------------------
\paragraph{Frequency heuristic.}
Following the frequency heuristic of \citet{shi2025trustnlp},
the LLM generates 10 answers at temperature 0.7
using the prompt in Table~\ref{tab:prompt_heuristic}.
The query is deemed ambiguous if the most frequent answer
accounts for less than a tuned threshold fraction of all responses.

\paragraph{Entropy heuristic.}
This baseline uses the same answer-oriented prompt and sampling procedure as the frequency heuristic
(Table~\ref{tab:prompt_heuristic}).
It computes the Shannon entropy of the sampled answer distribution and predicts ambiguity
when entropy exceeds a validation-tuned threshold.
High entropy indicates many distinct answers with similar frequencies;
low entropy indicates one dominant answer.

\paragraph{Intent-Sim~\cite{zhang2025clarify}.}
For each query, the LLM first generates a clarifying question
and then samples 10 hypothetical user responses at temperature 0.5.
We cluster the responses by pairwise entailment using DeBERTa-large fine-tuned on MNLI.
We classify queries as ambiguous when intent entropy exceeds a validation-tuned threshold.

\paragraph{APA (SFT)~\cite{kim2024apa}.}
This self-supervised fine-tuning baseline uses answer variability as a proxy for ambiguity.
For each query, the LLM generates candidate answers and clarification questions,
then fine-tunes on its own generated labels to predict ambiguity.

\paragraph{Shi et al. (RF)~\cite{shi2025trustnlp}.}
This feature-engineering baseline trains a random forest on diversity features from sampled LLM responses.
The LLM samples 10 responses across three temperatures: 3 at 0.3, 4 at 0.5, and 3 at 0.7.
The random forest uses lexical overlap, semantic similarity, and frequency statistics to classify ambiguity.

%--------------------------------------------------------------
% Context Quality Sensitivity
%--------------------------------------------------------------
\section{Context Quality Sensitivity}
\label{app:exp_context}
Table~\ref{tab:context} compares ambiguity detection on factoid queries with the LLM conditioned on no external context,
DPR-retrieved context~\cite{karpukhin2020densepassageretrievalopendomain},
and oracle context.
\method generally benefits from stronger grounding, achieving its highest scores under oracle context on both backbones.
It obtains the best \Famb in every setting and the best \Funamb in five of six settings.
The only exception is Qwen2.5-14B with DPR-retrieved context, where CoT w/ taxonomy surpasses \method on \Funamb (83.02 vs.\ 80.51).
Moreover, \method's \Funamb decreases from no context to DPR retrieval on both backbones
(LLaMA-2-7B: 85.57 $\rightarrow$ 77.48; Qwen2.5-14B: 85.66 $\rightarrow$ 80.51),
indicating that imperfect retrieval introduces noise for unambiguous queries.

%--------------------------------------------------------------
% Detailed NLI Relation Grids
%--------------------------------------------------------------
\section{Detailed NLI Relation Grids}
\label{app:nli_grids}
Figure~\ref{fig:nli_grid_detail} maps grid indices to generated answers
for three example queries.
Each cell is colored by the highest-probability NLI class:
green for entailment, blue for neutral, and red for contradiction.
The underlying $\tilde{S}$ retains the full distribution over
$\{\text{entailment}, \text{neutral}, \text{contradiction}\}$ for each cell.

\begin{figure*}[!t]
\centering

% --- Row 1: KDAmb (Georgia) ---
\begin{minipage}[c]{0.40\textwidth}
\centering
\includegraphics[width=\linewidth]{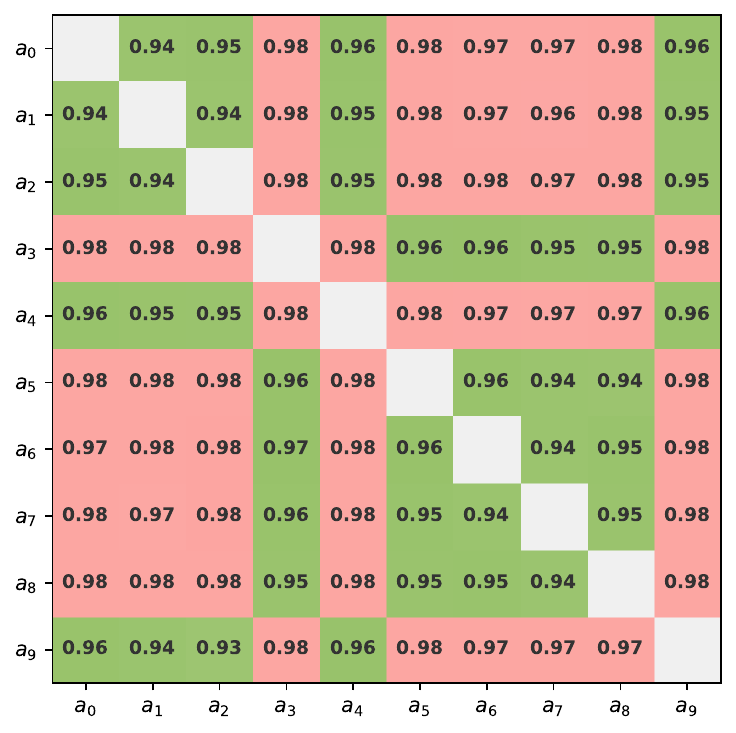}
\end{minipage}%
\hfill
\begin{minipage}[c]{0.55\textwidth}
\textbf{Knowledge-dependent ambiguous} \\
\textbf{Query:} What is the capital of Georgia?
\vspace{4pt}

{\normalsize
\begin{tabular}{@{}r@{\hspace{0.3em}}l@{}}
$a_0$: & Atlanta is the capital of Georgia. \\
$a_1$: & Atlanta serves as the capital of Georgia. \\
$a_2$: & The capital city of Georgia is Atlanta. \\
$a_3$: & Georgia's capital is Tbilisi. \\
$a_4$: & Georgia's capital is Atlanta. \\
$a_5$: & The capital of Georgia is Tbilisi. \\
$a_6$: & Tbilisi is the capital of Georgia. \\
$a_7$: & Tbilisi serves as the capital of Georgia. \\
$a_8$: & The capital city of Georgia is Tbilisi. \\
$a_9$: & The capital of Georgia is Atlanta. \\
\end{tabular}}
\end{minipage}

\vspace{12pt}

% --- Row 2: Closed (Rome) ---
\begin{minipage}[c]{0.40\textwidth}
\centering
\includegraphics[width=\linewidth]{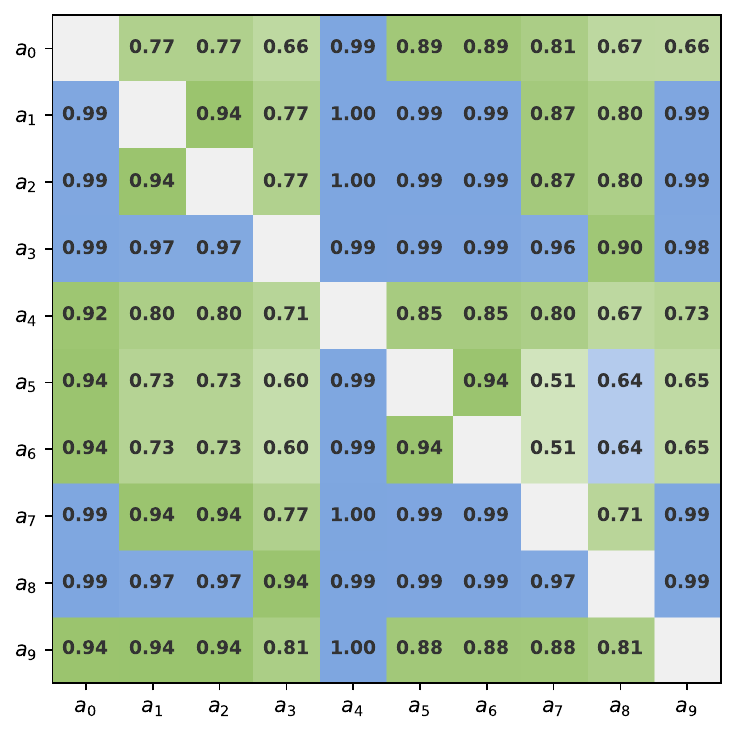}
\end{minipage}%
\hfill
\begin{minipage}[c]{0.55\textwidth}
\textbf{Closed unambiguous} \\
\textbf{Query:} Where is the Colosseum?
\vspace{4pt}

{\normalsize
\begin{tabular}{@{}r@{\hspace{0.3em}}l@{}}
$a_0$: & The Colosseum is in Rome. \\
$a_1$: & The Colosseum is in Italy. \\
$a_2$: & The Colosseum is in Italy. \\
$a_3$: & The Colosseum is in Europe. \\
$a_4$: & The Colosseum is in central Rome. \\
$a_5$: & The Colosseum is in the city of Rome. \\
$a_6$: & The Colosseum is in the city of Rome. \\
$a_7$: & The Colosseum is in the country of Italy. \\
$a_8$: & The Colosseum is in the continent of Europe. \\
$a_9$: & The Colosseum is in Rome, Italy. \\
\end{tabular}}
\end{minipage}

\vspace{12pt}

% --- Row 3: Open (Desk) ---
\begin{minipage}[c]{0.40\textwidth}
\centering
\includegraphics[width=\linewidth]{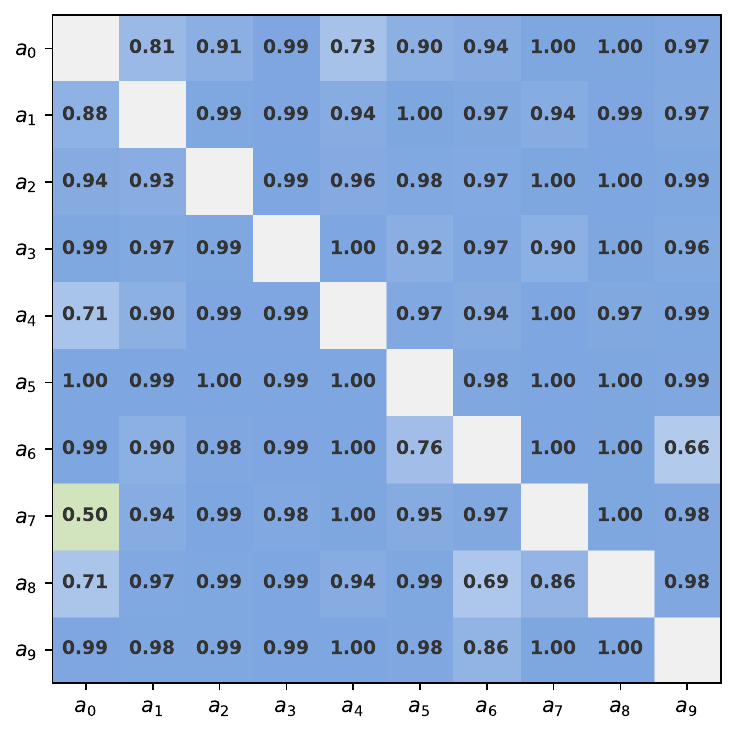}
\end{minipage}%
\hfill
\begin{minipage}[c]{0.55\textwidth}
\textbf{Open unambiguous} \\
\textbf{Query:} How to organize a desk?
\vspace{4pt}

{\normalsize
Each answer begins with ``A desk can be organized by''. \\[2pt]
\begin{tabular}{@{}r@{\hspace{0.3em}}l@{}}
$a_0$: & using storage containers. \\
$a_1$: & regularly cleaning the surface. \\
$a_2$: & using vertical space such as shelves or organizers. \\
$a_3$: & aligning items neatly to maintain visual order. \\
$a_4$: & keeping cables managed and untangled. \\
$a_5$: & removing unnecessary items to reduce clutter. \\
$a_6$: & grouping related items together. \\
$a_7$: & storing rarely used items away from the main surface. \\
$a_8$: & using a drawer organizer for small supplies. \\
$a_9$: & dedicating specific zones for different tasks. \\
\end{tabular}}
\end{minipage}

\vspace{4pt}
\caption{
\textbf{Refined relation grids reveal whether candidate answers conflict or differ.}
Cell $(i,j)$ is colored by the dominant NLI relation from $a_i$ to $a_j$: green denotes entailment, blue denotes neutral, and red denotes contradiction.
For knowledge-dependent ambiguity (top), competing candidate answers (Atlanta vs.\ Tbilisi) form a distributed contradiction pattern;
for closed unambiguous queries (middle), compatible answers at varying granularity mostly entail or remain neutral; and
for open unambiguous queries (bottom), diverse valid answers are largely neutral.
}
\label{fig:nli_grid_detail}
\end{figure*} 
\endgroup

\end{document}